\documentclass[lettersize,journal]{IEEEtran}
\usepackage{amsmath,amsfonts}
\usepackage{algorithmic}
\usepackage{algorithm}
\usepackage{array}
\usepackage[caption=false,font=normalsize,labelfont=sf,textfont=sf]{subfig}
\usepackage{textcomp}
\usepackage{stfloats}
\usepackage{url}
\usepackage{verbatim}
\usepackage{graphicx}
\usepackage{cite}
\usepackage{xcolor}
\usepackage{booktabs, multirow}       
\usepackage{bbding}
\usepackage{multirow}
\usepackage[hidelinks]{hyperref}
\begin{document}

\title{Practical Collaborative Perception: A Framework for Asynchronous and Multi-Agent \\ 3D Object Detection}

\author{Minh-Quan Dao$^1$, Julie Stephany Berrio$^2$, Vincent Frémont$^1$, Mao Shan$^2$, Elwan Héry$^1$, and Stewart Worrall$^2$ 
\thanks{This work has been supported by the French-Australian Science and Innovation Collaboration (FASIC) and the Australian Centre for Robotics (ACFR). The authors are with Ecole centrale de Nantes$^1$ (France) E-Mails: {\{minh-quan.dao, vincent.fremont, elwan.hery}\}@ls2n.fr and the ACFR at the University of Sydney$^2$ (Australia). E-mails: {$^2$ \{j.berrio, m.shan, s.worrall}\}@acfr.usyd.edu.au  }}

\maketitle

\begin{abstract}
Occlusion is a major challenge for LiDAR-based object detection methods as it renders regions of interest unobservable to the ego vehicle. 
A proposed solution to this problem comes from collaborative perception via Vehicle-to-Everything (V2X) communication, which leverages a diverse perspective thanks to the presence of connected agents (vehicles and intelligent roadside units) at multiple locations to form a complete scene representation.
The major challenge of V2X collaboration is the performance-bandwidth tradeoff which presents two questions (i) which information should be exchanged over the V2X network, and (ii) how the exchanged information is fused.
The current state-of-the-art resolves to the mid-collaboration approach where Birds-Eye View (BEV) images of point clouds are communicated to enable a deep interaction among connected agents while reducing bandwidth consumption.
While achieving strong performance, the real-world deployment of most mid-collaboration approaches are hindered by their overly complicated architectures and unrealistic assumptions about inter-agent synchronization.
In this work, we devise a simple yet effective collaboration method based on exchanging the outputs from each agent that achieves a better bandwidth-performance tradeoff while minimising the required changes to the single-vehicle detection models.
Moreover, we relax the assumptions used in existing state-of-the-art approaches about inter-agent synchronization to only require a common time reference among connected agents, which can be achieved in practice using GPS time. 
Experiments on the V2X-Sim dataset show that our collaboration method reaches 76.72 mean average precision which is 99\% the performance of the early collaboration method while consuming as much bandwidth as the late collaboration (0.01 MB on average).
The code will be released in \url{https://github.com/quan-dao/practical-collab-perception}.

\end{abstract}


\section{Introduction} \label{sec:intro}
The 3D detection and localization of objects is a fundamental module of an autonomous driving software stack. 
It provides input to downstream tasks such as object tracking, motion prediction, and navigation. Due to the requirement of localizing objects accurately in 3D, state-of-the-art methods use LiDARs as a primary sensing modality. 
While significant advancements have been made, occlusion remains troublesome for LiDAR-based detection models, which makes regions of interest inaccessible to the ego vehicle.
The issue of occlusion becomes particularly critical when navigating complex traffic scenarios, such as intersections, as autonomous vehicles are required to avoid collisions using a field of view that is severely reduced due to a large number of road users. 
The need for addressing occlusions at intersections is highlighted in a report published by Waymo \cite{schwall2020waymo} stating that two out of eight severe accidents involving their autonomous vehicles are due to occlusion in the intersections. 

Vehicle-to-Everything (V2X) collaborative perception is a promising solution to the occlusion challenge. 
Its core idea is to form a complete scene representation using measurements collected from multiple perspectives by leveraging the communication among multiple connected agents presenting at different locations. 
Connected agents can be either connected and automated vehicles (CAVs) or Intelligent Road-Side Units (IRSUs), which are advanced sensing systems strategically placed at elevated locations to have maximal coverage of regions where complex traffic scenarios take place.
This enhanced perception capacity, thanks to V2X, comes with several new technical challenges; the most notorious among them being the performance-bandwidth tradeoff which presents two questions; (i) which information should be broadcast, and (ii) how the exchanged information should be fused.

This tradeoff establishes a spectrum of solutions ranging from \textit{early} to \textit{late} collaboration. 
Raw measurements, which are point clouds in the context of this paper, are exchanged in the framework of \textit{early collaboration} to maximally reduce the impact of occlusion, thus achieving the highest performance at the expense of spending a significant amount of bandwidth. 
On the other extreme, \textit{late collaboration} exchanges high-level outputs (e.g., object detection as 3D bounding boxes) to minimize bandwidth usage, though this approach has been demonstrated to have reduced performance gains.
In an attempt to balance the two mutually excluding design targets, research on V2X collaboration frameworks \cite{wang2020v2vnet, li2021disconet, xu2022v2x-vit, xu2022bridging} are drawn toward the middle of this spectrum, thus the category's name of \textit{mid-collaboration}, 
where intermediate representations such as BEV images of agents' surrounding environment are chosen for broadcasting. 

While the motivation is just, most mid-collaboration methods require making substantial changes to the architecture of single-agent perception models to accommodate the fusion module where the combination of exchanged representations takes place. 
More importantly, these methods make strong assumptions about data synchronization among connected agents. 
For example, DiscoNet \cite{li2021disconet} considers a perfectly synced setting where agents share the same clock, collect and process point clouds at the same rate and the transmission/ receiving of BEV images experience zero latency. 
V2VNet \cite{wang2020v2vnet} and ViT-V2X \cite{xu2022v2x-vit} account for latency by postulating a global misalignment between exchanged BEV images and that of the ego vehicle. The cause of this misalignment is the movement of the ego vehicle between the time step it queries the V2X network and the time when the exchanged BEV images are received. 
This implicitly assumes that agents in the V2X network obtain point clouds synchronously.

Another drawback of the current state-of-the-art of V2X collaborative perception is that only one point cloud per agent is used.
Given that objects obscured in one frame may become visible in subsequent frames due to their movement or the motion of connected agents, and sparse regions might become dense as they draw nearer, harnessing sequences of point clouds is a compelling strategy to improve the performance of collaborative perception. 
In fact, the rich literature on multi-frame methods for single-vehicle object detection \cite{caesar2020nuscenes, hu2022afdetv2, yin2021center, djuric2021multixnet, laddha2021mvfusenet, tsai2023ms3d} has confirmed the effectiveness of point cloud sequences as a simple concatenation of point clouds in a common frame can boost detection accuracy by approximate $30\%$ \cite{caesar2020nuscenes}.

Finally, mid-collaboration methods assume connected agents share a common type of detector which is impractical for real-world deployment.
This assumption is critical because the domain gap between the BEV representation of point clouds made by different detectors severely affect their fusion \cite{xu2022bridging}.
Prior work \cite{xu2022bridging} resolves this challenge by introducing an additional module on top of those needed for the mid-collaboration to account for various differences in the BEV images made by SECOND \cite{yan2018second} and PointPillar \cite{luo2018fast}, thus further complicating the collaborative perception architecture. 

Aware of the aforementioned drawbacks of previous works on V2X collaborative perception, we seek a practical collaboration framework that emphasizes:
\begin{itemize}
    \item Minimal bandwidth consumption
    \item Minimal changes made to single-agent models
    \item Minimal inter-agent synchronization assumptions
    \item Support heterogeneous detectors networks
\end{itemize}
We aim our design at minimal bandwidth consumption, which can only be achieved by exchanging information about the objects detected by each agent. 
This design choice naturally satisfies the second design target as it dismisses the need for complex mid-representation fusion modules. 
We achieve the third target by only assuming that connected agents share a common time reference which is practically achievable using GPS time. 
To reach the last target, we decide to perform the fusion of exchanged information in the input of the ego vehicle's detector.

The challenge posed by our relaxed inter-agent synchronization assumption is that information (i.e., detected objects) broadcast by agents in the V2X network may never have the same timestamp as the query made by the ego vehicle. 
In other words, the detections made by other agents that are available on the V2X network always have an older timestamp compared to the current timestamp of the ego vehicle. 
This timestamp mismatch results in a misalignment between exchanged detected objects and their associated ground truths (if the detections are true positives), thus risking the overall performance. 
Our solution to this issue lies in the information that prior works have neglected - point cloud sequences.
We reason that objects detected in the past can be propagated to the present if their velocities are available.
The prediction of objects' velocities pertains to motion prediction \cite{phan2020covernet, djuric2021multixnet, varadarajan2022multipath++} or object tracking \cite{weng2020ab3dmot} which can negate our minimal architecture changes design target.
Since we assume agents produce predictions at least at the rate they acquire point clouds, we only need short-term (e.g., 0.1 seconds if point clouds are collected at 10Hz) velocity prediction, which can be computed using scene flow, rather than long-term prediction (e.g., 3 seconds \cite{phan2020covernet}) offered by motion prediction or tracking.
As a result, we use the plug-in module for scene flow estimation developed by our previous work \cite{dao2023aligning}.
This choice effectively makes our V2X collaboration framework a multi-frame method, thus enabling each connected agent, as an individual, to enjoy a boost in detection accuracy as single-vehicle multi-frame methods do.

This paper makes the following contributions:
\begin{itemize}
    \item Deriving a practical framework for V2X collaborative perception that outperforms \textit{Early Collaboration} while consuming as much bandwidth as \textit{Late Collaboration}. In addition, our method does not make any assumptions about inter-agent synchronization except the existence of a common time reference, introduces minimal changes to the architecture of single-vehicle detectors, and supports heterogeneous detector networks.
    \item Demonstrate the benefit of point cloud sequences in V2X collaborative perception
    \item Extending our previous work \cite{dao2023aligning} to further boost single-vehicle object detection accuracy and achieves more accurate scene flow prediction
    \item Performing extensive evaluations on NuScenes \cite{caesar2020nuscenes}, KITTI \cite{geiger2012kitti}, and V2X-Sim \cite{li2022v2x-sim} datasets to validate the performance our method
\end{itemize}

The remainder of this paper is organized as follows. 
Sec.II reviews the related works on V2X collaborative perception and highlights the gap to which we aim to contribute.
Next, our methodology is explained in Sec.III. 
Sec.IV presents our experimental results. 
Finally, the conclusion and future perspective are drawn in Sec.V.

\section{Related Work} \label{sec:related_works_v2x}

\begin{figure}[tb]
    \centering
    \includegraphics[width=0.95\linewidth]{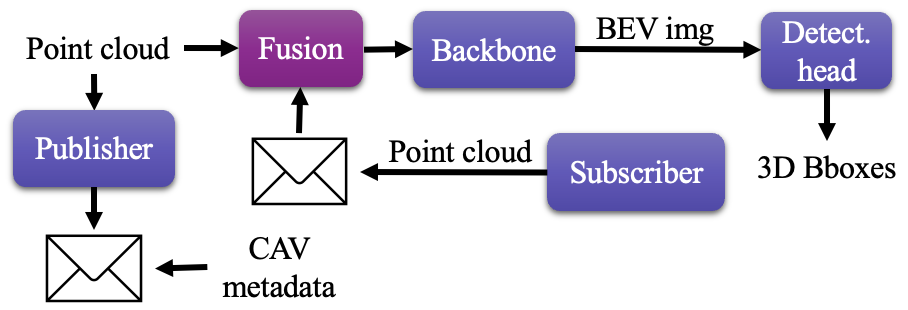}
    \caption{\small Early Collaboration framework. 
    Point clouds broadcast by connected agents are concatenated with that of the ego vehicle, forming the input to its detector.}
\label{fig:v2x_early}
\end{figure}

As described in the previous section, the main challenge of V2X cooperative perception is the performance-bandwidth tradeoff which establishes a solution spectrum ranging from \textit{early} to \textit{late} collaboration. 
In the \textit{early collaboration} approach, as depicted in Fig.\ref{fig:v2x_early}, agents exchange their raw measurements - point clouds. 
At every timestep, the ego vehicle concatenates its own point cloud with those obtained by other agents to form the input for its perception model. 
This combined point cloud offers a comprehensive view of the scene with minimal occlusion and sparsity thanks to the diverse perspectives of connected agents, thus being regarded as the upper bound of the performance of the V2X collaborative perception \cite{li2022v2x-sim, yu2022dair-v2x}. 
However, due to the significant amount of bandwidth required to transmit raw point clouds (the order of 10 MB), the early collaboration strategy is not feasible for real-world deployments.

\begin{figure}[b]
    \centering
    \includegraphics[width=0.96\linewidth]{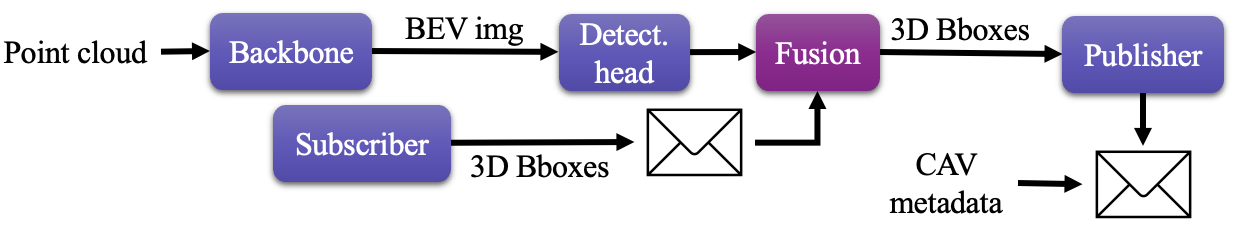}
    \caption{\small Late Collaboration framework.
    Objects detected by other agents are fused with those detected by the ego vehicle to form its final output.}
\label{fig:v2x_late}
\end{figure}

On the other extreme of the performance-bandwidth tradeoff, the \textit{late collaboration} approach focuses on minimizing bandwidth usage by exchanging only the detection results from each agent in the form of 3D bounding boxes. 
As illustrated in Fig.\ref{fig:v2x_late}, each agent independently detects objects using its own point cloud. 
Subsequently, the agent merges its own predictions with those made by others to generate the final output. 
While this strategy is more feasible for real-world deployments thanks to its minimal bandwidth consumption, it exhibits significantly lower performance gains compared to early fusion. 
The limited interaction among agents in the late collaboration approach contributes to this inferior performance. 
In noisy environments where latency is a factor, the late collaboration strategy even underperforms single-vehicle perception \cite{xu2022v2x-vit}.

\begin{figure}[h]
    \centering
    \includegraphics[width=0.96\linewidth]{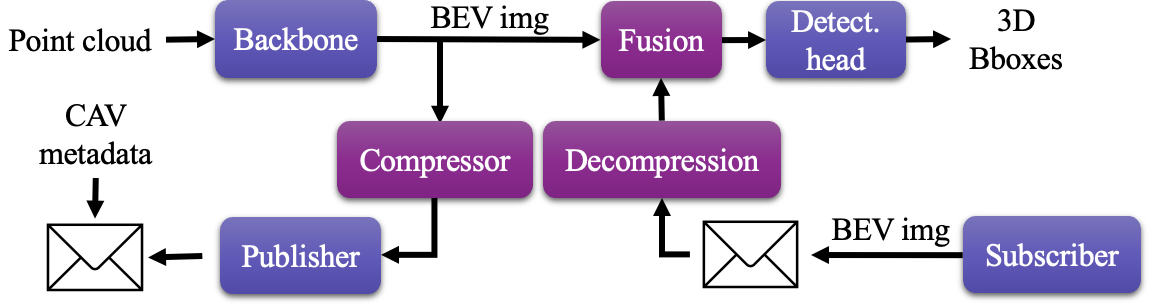}
    \caption{\small Mid Collaboration framework.
    The ego vehicle fuses the BEV image of its point cloud with those broadcast by other agents to improve its performance.}
\label{fig:v2x_mid_fusion}
\end{figure}

The \textit{mid-collaboration} strategy aims to find a balance between performance and bandwidth consumption by exchanging intermediate scene representations generated by the backbone of the agents' perception model. 
The motivation behind this approach is that the intermediate scene representation contains more contextual information compared to the final output (i.e., 3D bounding boxes), enabling deeper interaction among agents. 
Moreover, this representation is more compact than raw point clouds since it has been reduced in size through a series of convolution layers in the backbone and can be further compressed using an autoencoder to minimize bandwidth usage.
While the idea is elegant, implementing the intermediate collaboration strategy requires a range of modules, shown in Fig.\ref{fig:v2x_mid_fusion}, including compressor, decompressor, and representation fusion, among others, to match the performance of early collaboration. 
The fusion, in particular, is quite intricate as it involves learnable collaboration graphs using techniques such as Graph Neural Networks \cite{wang2020v2vnet, li2021disconet} or Transformers \cite{xu2022v2x-vit} to effectively fuse exchanged representations. 

In addition, dedicated modules are required to account for different practical challenges. 
For example, \cite{wang2020v2vnet, xu2022v2x-vit} use the Spatial Transformer \cite{jaderberg2015spatial} to resolve the global misalignment between the ego vehicle's representation and others' caused by the ego vehicle's motion between when it makes the query and when it receives exchanged information. 
In \cite{xu2022bridging}, the Fused Axial Attention \cite{xu2022cobevt} is used to bridge the domain gap between representations made by different detection models (e.g., PointPillar \cite{lang2019pointpillars} and SECOND\cite{yan2018second}) used by different agents in the V2X network. 
Finally, most mid-collaboration methods make strong assumptions about inter-agent synchronization which is either (i) perfect synchronization where exchanged representations always share the same timestamp \cite{li2021disconet} or (ii) synchronized point cloud acquisitions, meaning all agents obtain and process point clouds at the same rate and at the same time \cite{wang2020v2vnet, xu2022v2x-vit}.  
Because of these complexities, the real-world deployment of intermediate fusion remains challenging.

\section{Methodology}

\begin{figure}[tb]
    \centering
    \includegraphics[width=0.99\linewidth]{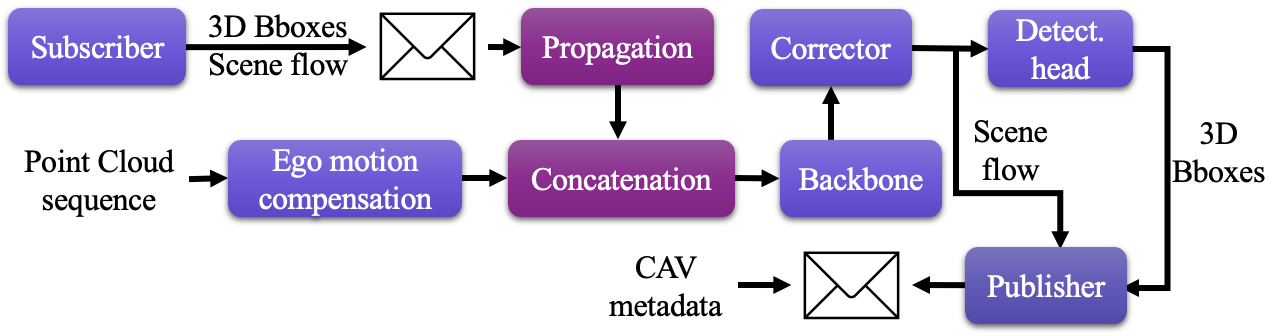}
    \caption{\small Our \textit{late-early} collaboration framework for V2X Cooperative Perception. 
    In which, objects detected by each connected agent at a past time that is closest to the present are broadcast.
    Exchanged detected objects are propagated to the present using their velocity computed by pooling point-wise scene flow, then fused with the point cloud which the ego vehicle collected at the present to enhance its perception.}
\label{fig:v2x_our_late_early_fusion}
\end{figure}

In this work, we aim to resolve the aforementioned complexities of mid-collaboration to obtain a practical framework for V2X collaborative perception. 
Our design is grounded in minimal bandwidth consumption as this is essential for real-world deployment at scale. 
To achieve this goal we choose object detection, in the form of 3D bounding boxes, as the information to be exchanged which is similar to the \textit{late collaboration} strategy. 
We further relax the assumption on inter-agent synchronization to agents sharing a common time reference (e.g., GPS time) and acknowledge that agents produce detections at different rates. 
As a result, exchanged detections always have older timestamps compared to the timestamp of the query made by the ego vehicle, thus risking a spatial misalignment between exchanged detections and their associated ground truth (if detections are true positive). 
We resolve this misalignment by predicting objects' velocity simultaneously with their locations by pooling point-wise scene flow which can be produced by integrating our \textit{Aligner} module \cite{dao2023aligning} to any BEV-based object detectors. 
Finally, we avoid the inferior performance of \textit{late collaboration} by devising a new collaboration strategy that fuses exchanged detections with the ego vehicle's raw point cloud for subsequent processing by its detection model. 
The resulting framework is illustrated in Fig.\ref{fig:v2x_our_late_early_fusion}.
We name our method \textit{late-early} collaboration as connected agents broadcast their outputs, which is the signature of \textit{late collaboration}, while the ego vehicle fuse received information with its point cloud, which is the signature of \textit{early collaboration}.

The innovation of our collaboration strategy lies in our recognition of the similarity between object detection using point cloud sequences and collaborative detection. 
In both cases, there is a need to fuse information obtained from diverse perspectives. 
Point cloud sequences involve capturing the motion of the ego vehicle, which results in varying viewpoints, while collaborative detection entails incorporating insights from other agents present in the environment. 
By drawing this parallel, we leverage the shared principle of fusing information from multiple perspectives to enhance the accuracy and robustness of both object detection approaches.
Specifically, we utilize the latest advancement in the multi-frame object detection literature, called \textit{MoDAR} \cite{li2023modar}, which interprets previously detected objects to 3D points with additional features made of object sizes, heading, confident score, and predicted class.
These points are propagated to the present and merged with the point cloud obtained at the present to form the input of any off-the-shelf detectors.

Since the scene flow plays a prominent role in our V2X collaboration framework by propagating past detections to the present time, this section first recalls our previous work \cite{dao2023aligning} and presents the extension we make to improve the accuracy of its scene flow prediction.
Next, our strategy for V2X collaborative perception that enables using asynchronous exchanged information is derived. 
 
\subsection{Learning Scene Flow} \label{sec:aligner++}

\begin{figure}[htb]
    \centering
        \includegraphics[trim={0 0 0 0.1cm},clip,width=.98\linewidth]{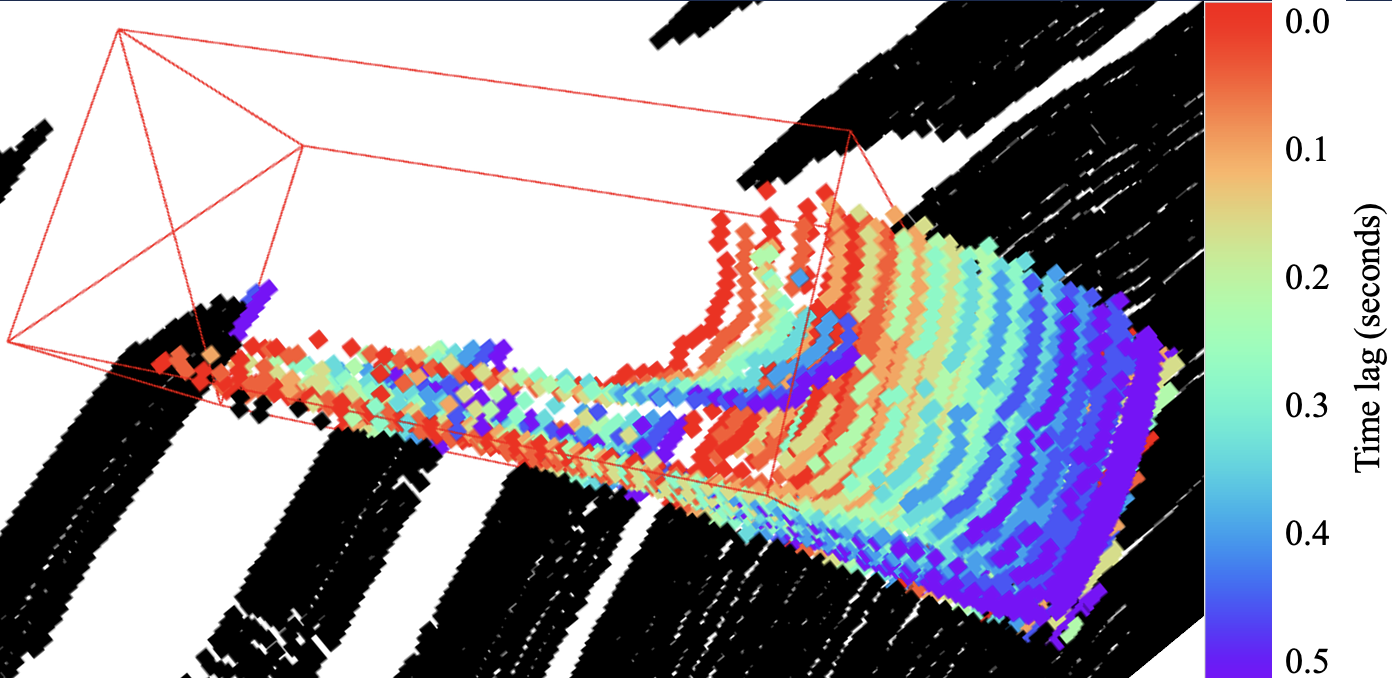}
\caption{\small A dynamic car's appearance on the concatenation of a 10-point-cloud sequence spanning 0.5 seconds. Points are color-coded according to their time lag with respect to the present.}
\label{fig:example_shadow_effect}
\end{figure}

As mentioned in Sec.\ref{sec:intro}, the accuracy of single-vehicle detection can be significantly improved by using concatenated point cloud sequences as inputs. 
However, such concatenation is done by the Ego Motion Compensation (EMC) method which only accounts for the motion of the ego vehicle resulting in the shadow effect, as can be seen in Fig.\ref{fig:example_shadow_effect}, in the concatenated point cloud. 
This effect, which is a misalignment in the 3D space between objects' points and their locations, leads to a misalignment in feature spaces (e.g., BEV representations), thus degrading detection accuracy for dynamic objects. 
In this section, we recall the approach we use in \cite{dao2023aligning} to handle this effect and present the extensions we make to it.
 
\subsubsection{The Cause of The Shadow Effect}
Let $\mathcal{P}^{t} = \left\{\mathbf{p}_j^t = [x, y, z, r, \Delta] ~ | ~ j = 1, \dots, N \right\}$ denote a point cloud collected by the ego vehicle at time step $t$. Here, each point $\mathbf{p}_j$ has two features namely intensity $r$ and time-lag $\Delta$ with respect to a predefined time step. The point cloud is expressed in the ego vehicle frame $\mathcal{E}(t)$ measured with respect to a global frame  $\mathcal{G}$. To concatenate a point cloud sequence $\mathcal{S} = \left\{\mathcal{P}^{t - K}, \mathcal{P}^{t - K + 1}, \dots,  \mathcal{P}^{t}\right\}$ of length $K + 1$, each point of point cloud $\mathbf{P}^{t - \Delta t}$ is transformed to the global frame using the ego vehicle pose at the same time step as in (\ref{eq:def_emc}), hence the method's name Ego Motion Compensation.
\begin{equation}
\label{eq:def_emc}
    ^{\mathcal{G}}\mathbf{p}_i^{t - \Delta t} =~ ^{\mathcal{G}}\mathbf{T}_{\mathcal{E}(t - \Delta t)} \cdot ~ \mathbf{p}_i^{t - \Delta t}
\end{equation}
Here, $^{d}\mathbf{T}_{s} \in \mathbf{SE(3)}$ represents the rigid transformation that maps points in frame $s$ to frame $d$. EMC undoes the motion of the ego vehicle without accounting for objects' motion. As a result, the appearance of a dynamic object in the concatenated point cloud comprises several instances, each corresponding to the object's poses in the global frame at a particular time step, as in Fig.\ref{fig:example_shadow_effect}.

Such distortion can be rectified by first transforming an object's points collected at different time steps to its body frame, and then from to the global frame at the desired time step. This two-step transformation is illustrated in (\ref{eq:def_rectification_tf}).
\begin{equation}
\label{eq:def_rectification_tf}
    ^{\mathcal{G}}\mathbf{\hat{p}}^{t - \Delta t} =~ ^{\mathcal{G}}\mathbf{T}_{\mathcal{O}(t)} \cdot ~ ^{\mathcal{O}(t - \Delta t)}\mathbf{T}_{\mathcal{G}} \cdot ~ ^{\mathcal{G}}\mathbf{p}^{t - \Delta t}
\end{equation}
Here, $\mathcal{O}(t)$ represents the object frame at time $t$. The point's subscript is omitted for clarity.

Equation (\ref{eq:def_rectification_tf}) implies that $^{\mathcal{O}(t)}\mathbf{T}_{\mathcal{O}(t - \Delta t)}$ is the identity transformation, which is the case for rigid objects. More importantly, the computation of the \textit{rectification transformation}, $^{\mathcal{G}}\mathbf{T}_{\mathcal{O}(t)} \cdot ~ ^{\mathcal{O}(t - \Delta t)}\mathbf{T}_{\mathcal{G}}$, requires object poses which are not available at test time. Our previous work \cite{dao2023aligning} develops a model that estimates such transformation directly from the concatenation of point cloud sequences. The following section briefly recalls the method used in \cite{dao2023aligning} and presents our extension.

\subsubsection{Learning to Rectify the Shadow Effect} \label{sec:aligner++}
\begin{figure*}[htb]
    \centering
    \includegraphics[width=0.98\linewidth]{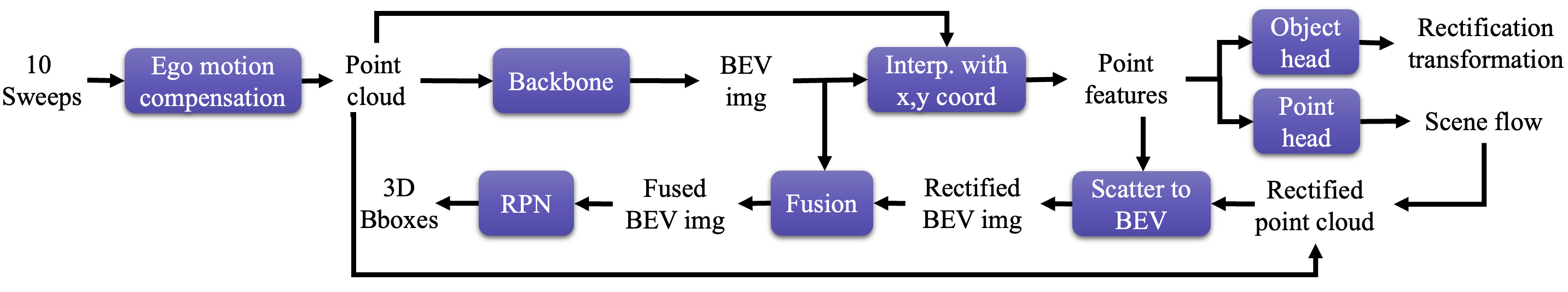}
    \caption{Aligner's overall pipeline. 
    Point-wise scene flow is predicted by Point Head based on point-wise features which are interpolated from the BEV image of the input point cloud.
    The input point cloud is then rectified by adding the corresponding scene flow to the coordinate of each LiDAR point.
    The scattering of point-wise features to the BEV using their rectified coordinate results in the rectified BEV image which is shadow-effect-free but sparse.
    The original BEV image which suffers from the shadow effect and the rectified BEV image are fused to minimize the shadow effect and sparsity. 
    The resulting fused BEV image is finally fed to the RPN to obtain object detection.
    }
\label{fig:arch_single_vehicle}
\end{figure*}

In \cite{dao2023aligning}, we propose a drop-in module, named \textit{Aligner}, for 3D object detection models that creates a BEV representation, $\mathbf{B} \in \mathbb{R}^{C \times H \times W}$, of the scene as an intermediate output. Our module first computes the feature of $\mathbf{f}_i \in \mathbb{R}^C$ of a point $\mathbf{p}_i$ by the bilinear interpolation of $\mathbf{B}$ using its projection to the BEV. Each point feature is then decoded into the point's scene flow $\mathbf{d}_i \in \mathbb{R}^3$ representing how much this point needs to offset to rectify the shadow effect. 
After being rectified, the new point cloud is used to scatter the set of point features $\left\{\mathbf{f}_i\right\}$ to the BEV to obtain a new BEV image $\mathbf{B}'$. While undergoing minimized impact of the shadow effect thanks to the rectification, $\mathbf{B}'$ is sparse as only occupied pillars have features, which has a negative effect on detection accuracy \cite{vedder2022sparse}. 
On the other hand, $\mathbf{B}$ is semi-dense thanks to the occupancy leakage caused by the convolutions in the detection models' backbone but possesses feature misalignment due to the shadow effect. To utilize the best of the two representations, they are fused before being fed to the Region Proposal Networks (RPN) for 3D bounding boxes. The complete pipeline is summarized in Fig.\ref{fig:arch_single_vehicle}.

The learning target $\mathbf{d}^*$ for the scene flow $\mathbf{d}$ of a point $\mathbf{p}^{t - \Delta t}$, which has the timestamp $(t - \Delta t)$ (i.e., collected at time step $t - \Delta t$), is the difference between its location and its rectified location computed by (\ref{eq:def_rectification_tf}).
\begin{equation}
    \mathbf{d}^* =~ ^{\mathcal{G}}\mathbf{\hat{p}}^{t - \Delta t} -  ~ ^{\mathcal{G}}\mathbf{p}^{t - \Delta t}
\end{equation}

A 3D bounding box is parameterized by a seven-vector comprised of its center coordinate $\left[c_x, c_y, c_z\right]$, its size $\left[l, w, h\right]$, and its heading $\theta$. The learning target is defined according to the framework which the RPN follows (e.g., anchor-based \cite{yan2018second} or center-based \cite{yin2021center}).

\subsubsection{Aligner++}
To improve the accuracy of \cite{dao2023aligning} in estimating scene flow, thus ultimately improving object detection, we introduce two extensions (i) incorporating HD Map and (ii) distilling a model trained on the concatenation of point cloud sequences by the ground truth trajectories. 

Previous works in integrating HD Maps into 3D object detection models \cite{yang2018hdnet, fang2021mapfusion} opt for a mid-fusion approach that concatenates rasterized maps with backbone-made BEV representations. 
However, this approach is incompatible with copy-paste data augmentation \cite{yan2018second}, which randomly samples ground truth objects from a database and pastes them to each point cloud used for training, as pasted objects do not necessarily adhere to the semantics and geometry of the map (e.g., cars appear inside a building). 
Moreover, they only utilize the map's binary channels such as drivable areas, sidewalks, or car parks while omitting the lane direction which is potentially helpful for estimating scene flow and objects' heading direction. 

\begin{figure}[tb]
    \centering
    \includegraphics[trim={0 5cm 0 5cm},clip, width=0.96\linewidth]{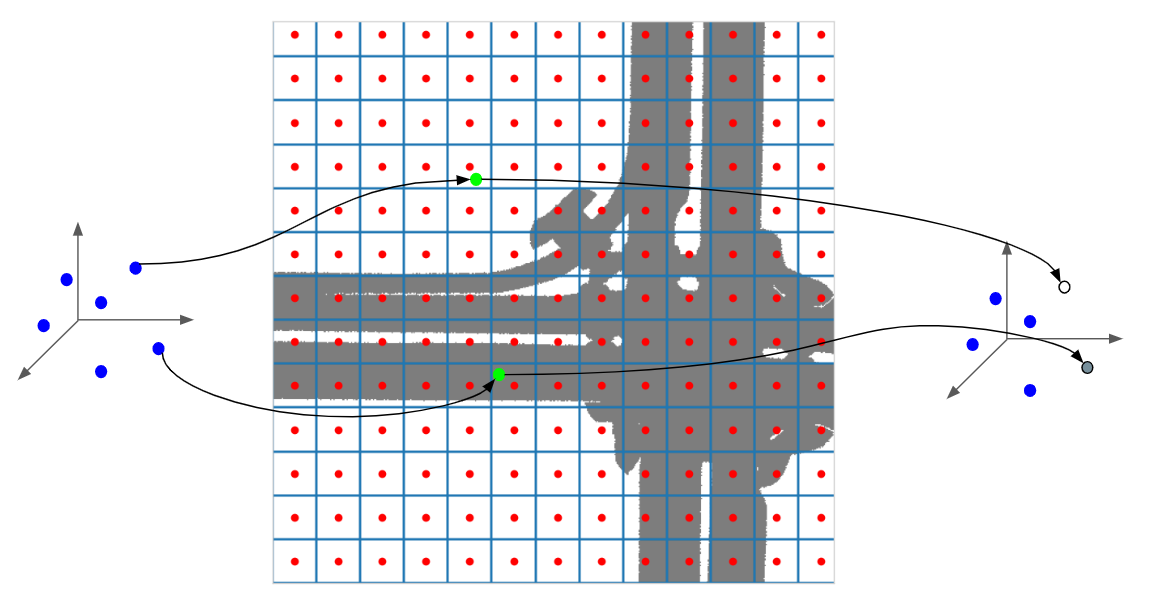}
    \caption{Extraction of map features.
    LiDAR points are projected to the BEV.
    Their map feature is obtained by nearest-neighbor interpolation on the rasterized HD Map.}
\label{fig:extact_map_feat_for_point}
\end{figure}

Aware of these two limitations, we propose to extract the map feature for each 3D point and use it to augment the point's raw features prior to further processing (e.g., building a ground truth database or computing a BEV representation for object detection). 
The map feature extraction process is illustrated in Fig.\ref{fig:extact_map_feat_for_point} where points' coordinate in the map's BEV is used for nearest neighbor interpolation. 
This attachment of the map features to points results in an early fusion between HD Maps and LiDARs, rendering the concatenation of maps' channels to backbone-made BEV representations unnecessary. 
As a result, HD Maps are made compatible with copy-paste data augmentation.

\begin{figure*}
    \centering
    \vspace{2mm}
    \includegraphics[width=0.98\linewidth]{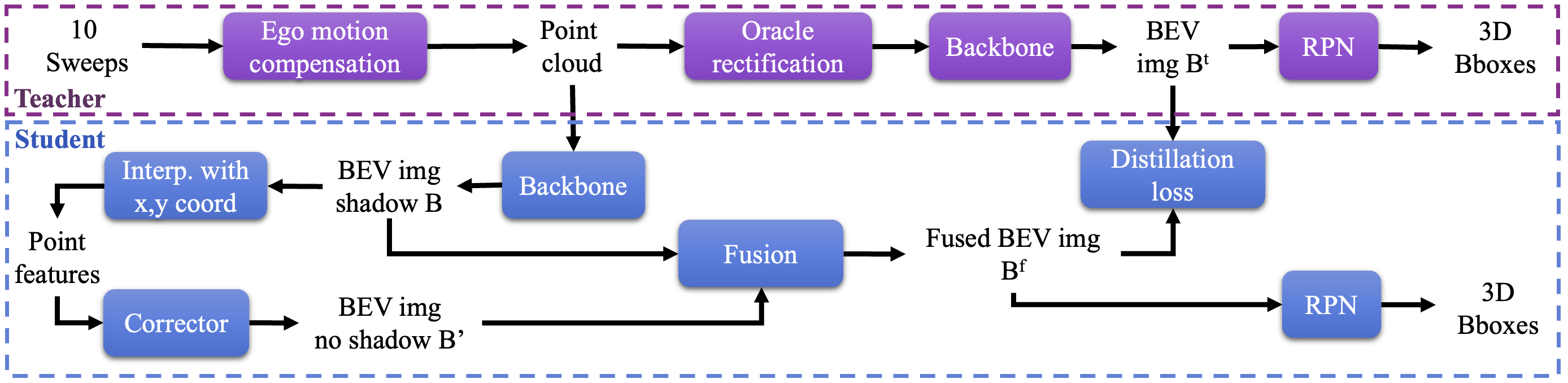}
    \caption{\small Alinger++ 's teacher-student framework.
    A model, pre-trained on point clouds concatenated using object ground truth trajectories, plays the role of the teacher.
    During training, the student model, which operates on point clouds concatenated using ego vehicle localization, strives to emulate the BEV images made by the teacher by minimizing a distillation loss.}
\label{fig:teacher_student}
\end{figure*}

\begin{figure}[htb]
\centering
\subfloat[]{
    \includegraphics[trim={0 10cm 0 6cm},clip, width=.95\linewidth]{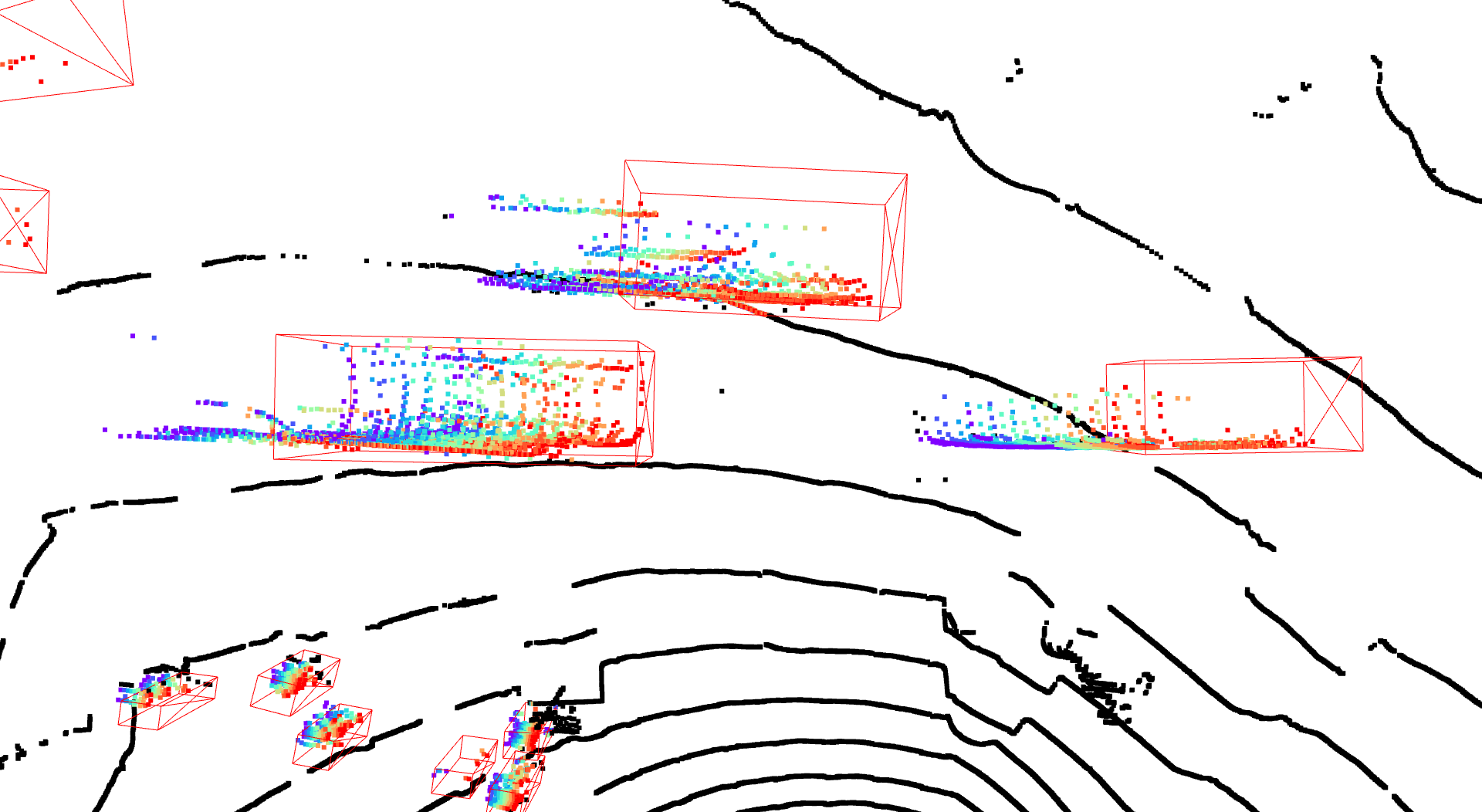}
    \label{fig:raw_sequence_shadow}
} \\
\vfill
\subfloat[]{
    \includegraphics[trim={0 10cm 0 6cm},clip,width=.95\linewidth]{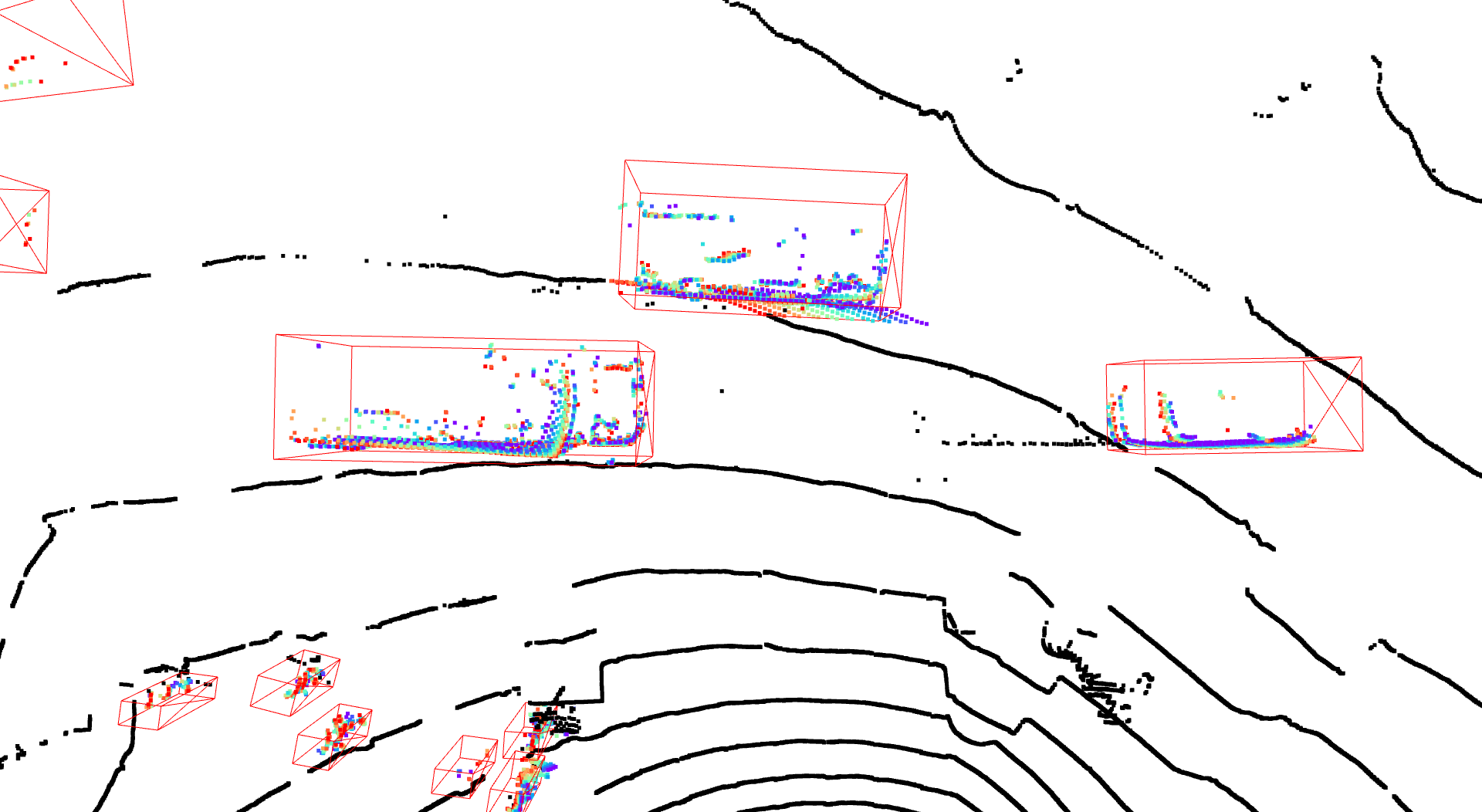}%
    \label{fig:raw_sequence_corrected}
}
\vfill
\subfloat[]{
    \includegraphics[trim={4.3cm 0 3.3cm 1.3cm},clip,width=0.96\linewidth]{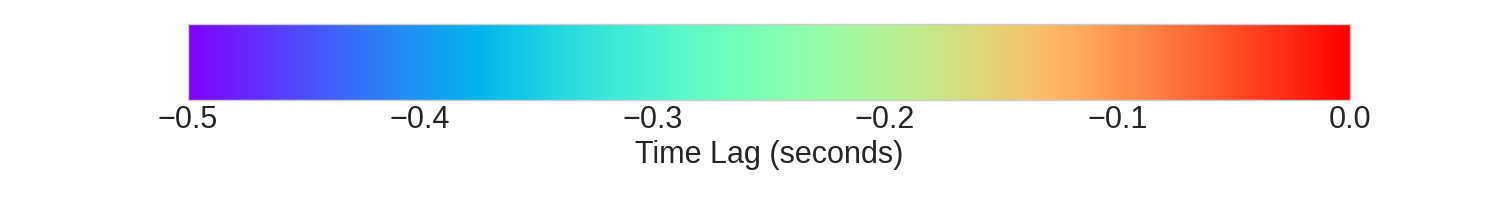}%
}
\caption{Comparison between concatenation of a point cloud sequence (a) and its rectification using objects' ground truth trajectories (b). Points are color-coded according to their time lag with respect to the present. The color bar is shown in (c).}
\label{fig:point_cloud_for_teacher}
\end{figure}

To improve the quality of the fused BEV representation, $\mathbf{B}^f$ in Fig.\ref{fig:arch_single_vehicle}, in terms of minimizing the feature misalignment caused by the shadow effect and reducing the sparsity, we use the teacher-student framework shown in Fig.\ref{fig:teacher_student}. 
In detail, we use ground truth objects' trajectories to rectify the concatenation of point cloud sequences and use the shadow-effect-free results, illustrated in Fig.\ref{fig:point_cloud_for_teacher}, to train a teacher model. After the teacher converges, its BEV representation $\mathbf{B}^t$ is used to guide the student's fused BEV representation $\mathbf{B}^f$ by optimizing the students' weight so that the difference measured by the $\mathit{L}_2$ loss between these two representations is minimized.

\subsection{V2X Collaborative Perception Framework}

The V2X setting that we target in this paper comprises multiple CAVs and IRSUs which are collectively referred to as agents. 
An agent $\mathcal{A}_i$ is equipped with a LiDAR to localize in a common global frame $\mathcal{G}$ and detect objects in its surrounding environment. 
The detection is based on the processing of point cloud sequences using a detection model made of the integration of the \textit{Aligner++} presented in the Sec.\ref{sec:aligner++} into an off-the-shelf single-frame object detector such as PointPillar \cite{lang2019pointpillars}. 
At a time step $t_i$, agent $\mathcal{A}_i$ uses a $K$-point-cloud sequence $\mathcal{S}_i = \left\{\mathcal{P}^{t_i - K + 1}_i, \dots, \mathcal{P}^{t_i}_i\right\}$ as an input to its detection model. 
An object $\mathbf{b}_{i, j}$ detected by agent $\mathcal{A}_i$ is parameterized by a nine-vector $[x, y, z, w, l, h, \theta, s, c]$. 
The first seven numbers localize the object by its center location $[x, y, z]$, size $[w, l, h]$, heading direction $\theta$. 
The last two numbers, $s$ and $c$, respectively denote confidence score $s$ and the predicted class $c$.

Upon receiving a query having timestamp $t$, agent $\mathcal{A}_i$ will communicate its detection $\mathcal{B}_i^{t_i} = \left\{\mathbf{b}_{i,j}\right\}_{j=1}^{M_i}$ and metadata produced timestamp $t_i$ that is prior to and closest to $t$. 
The agent's metadata produced at a timestamp $t_i$ is made of the timestamp itself and the agent's pose $\mathcal{E}_i(t_i)$ at this timestamp. 

Given this setting, we aim to enhance the ego vehicle's capacity of detecting objects by fusing its point cloud sequence $\mathcal{S}_e = \left\{\mathcal{P}^{t - K + 1}_e, \dots, \mathcal{P}^{t}_e\right\}$ with the \textit{MoDAR} interpretation of predictions $\mathcal{B}_i^{t_i}$ made by other agents. 
The following section provides an overview of the origin of \textit{MoDAR} and presents in detail how we adapt this concept to collaborative perception via the V2X context.

\subsubsection{MoDAR for Object Detection on Point Cloud Sequences} 
\textit{MoDAR} is created to enable detecting objects in extremely long point cloud sequences (hundreds of frames). 
In the \textit{MoDAR} framework, a sequence is divided into several short sequences where objects are detected by a single-frame detector, tracked by a simple multi-object tracker (e.g., \cite{weng2020ab3dmot}). 
Then, the data-driven motion forecasting model MultiPath++\cite{varadarajan2022multipath++} predicts objects' future poses based on objects' trajectories established by the tracker. 
Using prediction about objects' future poses, detected objects in each short subsequence are propagated to the desired time step (e.g., the present). 
Next, each propagated object, which is represented as an up-right 3D bounding box (parameterized by the location of its center, size, and heading) with a confidence score and a class, is converted into a 3D point that takes the box's center as its coordinate and the box's size, heading, confidence score and class as features. 
These points are referred to as \textit{MoDAR} points. 
They enable packing an entire subsequence into a small number of points, thus enabling an efficient fusion of extremely long point cloud sequences.

\subsubsection{V2X Collaboration using MoDAR Points} \label{sec:modar_for_v2x}
We draw the following similarity between single-vehicle object detection on point cloud sequences and V2X collaborative detection: these two tasks share the common challenge of finding an effective method for fusing information obtained from different perspectives caused by the motion of the ego vehicle in the case of point cloud sequences and the presence of other agents in the case of collaboration via V2X. Based on this observation, we use \textit{MoDAR} points as the medium for conveying information among agents in the V2X network. Specifically, we interpret an object detected by agent $\mathcal{A}_i$ as a 3D bounding box $\mathbf{b}_{i,j} = [x, y, z, w, l, h, \theta, s, c]$ to a \textit{MoDAR} point $\mathbf{m}_{i,j}$ by assigning
\begin{itemize}
    \item $[x, y, z]$ to the coordinate of $\mathbf{m}_{i,j}$
    \item $[ w, l, h, \theta, s, c]$ to $\mathbf{m}_{i,j}$'s features 
\end{itemize}

The challenge in our V2X setting is that different agents detect objects at different rates, thus forcing the ego vehicle to utilize \textit{MoDAR} points made by other agents at passed time steps. This timestamp mismatch results in a spatial misalignment between exchanged \textit{MoDAR} points and ground truth dynamic objects, which can diminish the benefit of collaboration or even decrease the ego vehicle's accuracy. This challenge is encountered in the context of single-vehicle detection on point cloud sequences as well because dynamic objects change their poses from one subsequence to another. \cite{li2023modar} resolves this by predicting objects' pose using a multi-object tracker and the motion forecasting model MultiPath++. 

While this is feasible in the V2X context, the implementation of those modules for future pose prediction does not align with our design target of minimal architecture. 
Instead, we use the scene flow to propagate \textit{MoDAR} points from a prior timestep to the timestep queried by the ego vehicle. 
Since \textit{MoDAR} points are virtual, their scene flow is not estimated directly from a point cloud sequence but is aggregated from the scene flow of points residing in the box they represent. 
Specifically, let $\mathbf{m}_{i, j}$ be a \textit{MoDAR} point representing a 3D bounding box $\mathbf{b}_{i, j}$ detected by agent $\mathcal{A}_i$ at timestep $t_i$. $\mathbf{p}_{i, h}$ and $\mathbf{f}_{i, h}$ respectively denote a real 3D point in the concatenation of agent $\mathcal{A}_i$'s point cloud sequence $\mathcal{S}_{i}$ and its predicted scene flow. 
The scene flow $\mathbf{f}^{\mathbf{m}}_{i, j}$ of $\mathbf{m}_{i, j}$ is computed by
\begin{equation}
    \mathbf{f}^{\mathbf{m}}_{i, j} = \mathrm{mean} \left\{\mathbf{f}_{i, h} ~ | ~ \mathbf{p}_{i, h} \in \mathbf{b}_{i, j} \right\}
\end{equation}
Once its scene flow is obtained, the \textit{MoDAR} point $\mathbf{m}_{i, j}$ is propagated to the timestep $t$ queried by the ego vehicle as following
\begin{equation}
    [\hat{\mathbf{m}}_{i, j}]_{x,y,z} = [\mathbf{m}_{i, j}]_{x,y,z} + \frac{t - t_i}{|\mathcal{S}_{i}|} ~ \mathbf{f}^{\mathbf{m}}_{i, j}
\label{eq:propagate_modar}
\end{equation}
Here, $|\mathcal{S}_{i}|$ denotes the length of the point cloud sequence $\mathcal{S}_{i}$ measured in seconds. 
$[\cdot]_{x,y,z}$ is the operator that extracts 3D coordinate of a \textit{MoDAR} point.

Finally, propagated \textit{MoDAR} point is transformed from the agent $\mathcal{A}_i$'s pose $\mathcal{E}_i(t_i)$ at time step $t_i$ to the ego vehicle pose $\mathcal{E}_e(t)$ at time step $t$ using the localization in the common global frame $\mathcal{G}$ of the two agents
\begin{equation}
    ^{\mathcal{E}_e(t)}[\hat{\mathbf{m}}_{i, j}]_{x,y,z} =~ ^{\mathcal{G}}\mathbf{T}^{-1}_{\mathcal{E}_e(t)} ~ ^{\mathcal{G}}\mathbf{T}_{\mathcal{E}_i(t_i)} ~ [\hat{\mathbf{m}}_{i, j}]_{x,y,z}
\end{equation}

The concatenation between the set of \textit{MoDAR} points received from other agents and the ego vehicle's raw point cloud (resulting from concatenating its own point cloud sequence) is done straightforwardly by padding
\begin{itemize}
    \item points in the ego vehicle's raw point cloud with null vectors representing features of \textit{MoDAR} points, which are boxes' size, heading, score, and class
    \item \textit{MoDAR} points with null vectors representing features of points in the ego vehicle's raw point cloud which are points' intensity and time-lag.
\end{itemize}
Once exchanged \textit{MoDAR} points and the point cloud of the ego vehicle are merged, the result can be processed by the single-vehicle model developed in the Sec.\ref{sec:aligner++} without changing its architecture.

\section{Experiments and Results}
This section first shows the results of the improvements to single vehicle perception with scene flow using real-world datasets, then demonstrates the collaborative approach using a simulated V2X dataset.

\subsection{Single-Vehicle Perception}

\subsubsection{Datasets and Metrics}
As point clouds' resolution has a large impact on the performance of LiDAR-based models,  we test our model on three resolutions: 16, 32, and 64. While 32-channel and 64-channel point clouds are readily available in NuScenes dataset \cite{caesar2020nuscenes} and KITTI dataset \cite{geiger2012kitti} respectively, to the best of our knowledge, there are no publicly available datasets containing 16-channel point clouds. As a result, we synthesize a 16-channel dataset from the NuScenes dataset using the downsampling approach of \cite{wei2022lidar}. 

\paragraph{The NuScenes dataset} is made of 850 20-second scenes split into 700 scenes for training and 150 scenes for validation. Each scene comprises data samples collected by a multimodal sensor suite including a 32-beam LiDAR operating at 20 Hz. In \textit{NuScenes}' convention, a keyframe is established once all sensors are in sync which happens every half a second. For each keyframe, objects are annotated as 3D bounding boxes. In addition, each object is assigned a unique ID that is kept consistent throughout a scene which enables us to generate the ground truth for the \textit{rectification transformation} and scene flow.
The downsampling of each point cloud in the \textit{NuScenes} dataset to synthesize a 16-channel one is done by first identifying its points' beam index via K-mean (K is 32 in the case of \textit{NuScenes}) clustering on their azimuth coordinate, then assigning an equivalent beam index with respect to 16-channel LiDAR. More details can be found in \cite{wei2022lidar}. Hereon, we refer to this synthesized dataset as \textit{NuScenes}-16.

\paragraph{The KITTI dataset} has a 3D Object Detection partition containing 7481 and 7581 samples for training and testing. Each sample comprises sensor measurements collected by a 64-beam LiDAR operating at 10Hz and several cameras. A common practice when working with \textit{KITTI} is to split the original training data into 3712 training samples and 3769 validation samples for experimental studies. A challenge we encounter when using \textit{KITTI} is that its 3D Object Detection partition contains temporally disjointed samples, thus being not straightforward to obtain input (point cloud sequences) and ground truth (scene flow) for our model. We resolve this challenge by matching samples in the 3D Object Detection partition with \textit{KITTI}'s raw sequences. Since there are a few raw sequences that do not have \textit{tracklet} annotation, meaning objects are not tracked, we only retain data samples whose associated raw sequences have \textit{tracklets} (i.e., trajectories of objects) in the training set and validation set.

\paragraph{Metrics} \label{sec:metrics}
We use mean Average Precision (mAP) to measure the performance of our model on the 3D object detection task. A prediction is matched with the closest ground truth, measured by an affinity. A match is considered valid if the affinity is below a predefined threshold. For each threshold, the average precision is obtained by integrating the recall-precision curve for recall and precision above 0.1. The mAP is the mean of the average precision of the threshold set. For \textit{NuScenes}, the affinity is the Euclidean distance on the ground plane between centers of predictions and ground truth. This distance has four thresholds: 0.5, 1.0, 2.0, and 4.0. For \textit{KITTI}, the affinity is the Intersection-over-Union (IoU) in the BEV plane. Unlike \textit{NuScenes}, \textit{KITTI} uses only one affinity threshold for each class, which is 0.7 for cars and 0.5 for pedestrians.

The evaluation of scene flow prediction is based on the set of standard metrics proposed by \cite{liu2019flownet3d} which includes End-Point Error (EPE), strict/ relaxed accuracy (AccS/ AccR), and outlier (ROutliers). The EPE is the Euclidean distance between the predicted scene flow and their ground truth average over the total number of points. The AccS/ AccR is the percentage of points having either EPE $<$ 0.05/ 0.10 meters or relative error $<$ 0.05/ 0.10. The ROutliers is the percentage of points whose EPE $>$ 0.30 meters and relative error $>$ 0.30.

\subsubsection{Experiments and Results} \label{sec:single_vehicle_implementation_detail}
The implementation of the single-vehicle perception model here follows the implementation made in our previous work \cite{dao2023aligning} which uses the concatenation of 0.5-second point cloud sequences by EMC as the input. Since \textit{NuScenes} and \textit{KITTI} respectively obtain point clouds at 20 and 10 Hz, a 0.5-second sequence contains 10 and 5 point clouds. We use the same architecture, which uses PointPillar \cite{lang2019pointpillars} as its backbone and CenterHead \cite{yin2021center} as its RPN, for every experiment. In experiments on \textit{NuScenes}, the detection range is limited to $[-51.2, 51.2]$ on the XY plane and $[-5.0, 3.0]$ along the Z axis. For experiments in \textit{KITTI}, point clouds are first cropped using the left-color camera's field of view as the annotations are only provided for objects that lie within the left-color camera's image. The detection range on \textit{KITTI} dataset is $[0, 69.12] \times [-39.68, 39.68] \times [-3, 1]$ along the X, Y, and Z axis. Our implementation is based on the OpenPCDet \cite{openpcdet2020}. The details on the model's hyperparameters can be found in our code release.

\paragraph{Scene Flow Evaluation}
The evaluation of the \textit{Aligner++} on scene flow estimation is done using the \textit{NuScenes} dataset. 
The result shown in Tab.\ref{tab:performance_flow3d} indicates a significant improvement compared to our previous work \cite{dao2023aligning} that reaches state-of-the-art on accuracy-related metrics namely AccS, AccR, and ROutliers while maintaining low inference time. 
The runtime of our methods, \textit{Aligner} and \textit{Aligner++}, is measured on an NVIDIA A6000 GPU, while the runtime of five baselines is measured on an NVIDIA RTX 3090 GPU (as reported by \cite{huang2022dynamic}) having the same compute capability.
The \textit{Aligner++} has relatively high EPE; this is because the evaluation is done for every point in the point cloud. 
This means the evaluation is carried out for both ground truth foreground points which have associated ground truth scene flow (can be nonzero or zero depending on dynamic) and ground truth background points which we assign the null vector as ground truth scene flow. 
Since the classification module of the Object Head inevitably makes false positive/ false negative foreground predictions, a number of background/ foreground points are predicted to have nonzero/ zero scene flow. 
Even though the portion of false predictions is small, as indicated by accuracy (AccS, AccR) and outlier metrics, the magnitude of their error is sufficiently large, due to the fact either the prediction or ground truth is zero, thus resulting in a large EPE.

\begin{table}[t]
    \caption{performance of our model on scene flow metrics}
    \centering
    \resizebox{\linewidth}{!}
    {
    \begin{tabular}{l c c c c r}
    \toprule
    \multirow{2}{*}{Method} & \multirow{2}{*}{EPE $\downarrow$} & \multirow{2}{*}{AccS $\uparrow$} & \multirow{2}{*}{AccR $\uparrow$} & \multirow{2}{*}{ROutliers $\downarrow$} & Runtime $\downarrow$ \\
                            &                                   &                                  &                                  &                                         & (seconds) \\
    \midrule
    FLOT \cite{puy2020flot} & 1.216 & 3.0 & 10.3 & 63.9 & 2.01 \\
    
    NSFPrior \cite{li2021neural} & 0.707 & 19.3 & 37.8 & 32.0 & 63.46 \\
    
    PPWC-Net \cite{wu2020pointpwc} & 0.661 & 7.6 & 24.2 & 31.9 & 0.99 \\
    
    WsRSF \cite{gojcic2021weakly} & 0.539 & 17.9 & 37.4 & 22.9 & 1.46 \\
    
    PCAccumulation \cite{huang2022dynamic} & 0.301 & \underline{26.6} & \underline{53.4} & \underline{12.1} & 0.25\\
    \midrule
    PointPillar + & \multirow{2}{*}{\underline{0.506}} & \multirow{2}{*}{16.8} & \multirow{2}{*}{30.2} & \multirow{2}{*}{33.8} & \multirow{2}{*}{\textbf{0.09}} \\
    Aligner \cite{dao2023aligning} &                   &                       &                       &                       & \\
    
    PointPillar + & \multirow{2}{*}{0.616} & \multirow{2}{*}{\textbf{46.4}} & \multirow{2}{*}{\textbf{66.6}} & \multirow{2}{*}{\textbf{6.8}} & \multirow{2}{*}{\underline{0.10}} \\
    Aligner++ &                   &                       &                       &                       & \\
    
    \bottomrule
    \end{tabular}
    }
\label{tab:performance_flow3d}
\end{table}

\paragraph{Object Detection Evaluation}
The better scene flow estimation made by the results in a higher detection accuracy on the \textit{NuScenes} dataset which can be seen in Tab.\ref{tab:performance_effect_of_extension}. The integration of \textit{Aligner++}, made using the HD Map and through distillation using the teacher-student framework, improves the mAP average over 10 classes of objects of a plain PointPillar by 6 points which is almost double the gain brought by \textit{Aligner} (3.4 points). Interestingly, the distillation is responsible for most (5.6 out of 6 points) of the success of the \textit{Aligner++}.

\begin{table}[b]
    \caption{detection improvement due to our extension}
    \centering
    \begin{tabular}{c c c c c}
    \toprule
    \multirow{2}{*}{PointPillar} & \multirow{2}{*}{Aligner \cite{dao2023aligning}} & \multirow{2}{*}{HD Map} & Distilling & mAP (avg  \\
                                 &                                                 &                         & Teacher    & 10 classes) \\
    \midrule
    \Checkmark &                                &               &               & 39.0 \\
    \Checkmark & \Checkmark                     &               &               & 42.4 \\
    \Checkmark & \Checkmark                              & \Checkmark    &               & 42.8 \\
    \Checkmark & \Checkmark                               & \Checkmark    & \Checkmark    & 45.0 \\
    \bottomrule
    \end{tabular}
\label{tab:performance_effect_of_extension}
\end{table}

The robustness of our \textit{Aligner++} is demonstrated by experiments on the \textit{KITTI} and the synthetic \textit{NuScenes-16} dataset. As can be seen Tab.\ref{tab:peformance_kitti_tempo} and Tab.\ref{tab:peformance_nuscenes_16}, the integration of \textit{Aligner++} consistently improves the accuracy of detecting objects in point cloud sequences. A common point between these two tables is that performance gain from \textit{Aligner++} is larger for vehicle-like classes (e.g., car, truck, or trailer). This is because the scene flow ground truth is generated using the \textit{rectification transformation} (\ref{eq:def_rectification_tf}) which is established based on the rigid motion assumption. This assumption holds for vehicle-like classes while being an oversimplification for pedestrians. As a result, the estimation of scene flow for vehicle-like classes is more accurate, thus resulting in higher detection accuracy. Another important result that can be drawn from Tab.\ref{tab:peformance_kitti_tempo} and Tab.\ref{tab:peformance_nuscenes_16} is the effectiveness, in terms of detection accuracy, of using point cloud sequences over single point clouds persists across various LiDAR resolutions.

\begin{table}[htb]
    \caption{detection performance on kitti dataset}
    \centering
    \begin{tabular}{c c c c c c}
    \toprule
    \multirow{2}{*}{Class} & \multirow{2}{*}{Model} & \multicolumn{4}{c}{Sequence Length} \\
                                                    \cmidrule{3-6}
                           &                        & 1      & 2     & 3     & 4 \\
    \midrule
    \multirow{2}{*}{Car}   & PointPillar            & 66.54  & 70.44 & 69.66  & 69.88 \\
                           & + Aligner++            & /      & +3.91 & +7.45  & +6.3 \\
    \midrule
    Pedes-                 & PointPillar            & 5.32  & 6.16  & 16.71  & 20.77 \\
    trian                  & + Aligner++            & /     & +4.82 & +4.07  & +3.6 \\
    \bottomrule
    \end{tabular}
\label{tab:peformance_kitti_tempo}
\end{table} 
 
\begin{table}[htb]
    \caption{detection performance on nuscenes-16}
    \centering
    \begin{tabular}{l c c  r}
    \toprule
    \multirow{2}{*}{Model}   & \multicolumn{2}{c}{\multirow{2}{*}{PointPillar}} & PointPillar \\
                             &                &                                 & + Aligner++ \\
    \midrule
    Seq Length        & 1         & 10         & 10     \\ 
    \midrule
    Car               & 56.14     & 79.84      & 80.81  \\
    Truck             & 26.99     & 48.64      & 49.39  \\
    Const. Vehicle    & 0.61      & 8.21       & 9.45   \\
    Bus               & 38.06     & 59.91      & 59.99  \\
    Trailer           & 11.05     & 26.15      & 29.03  \\
    Barrier           & 38.02     & 59.07      & 59.26  \\
    Motorcycle        & 12.70     & 41.14      & 41.72  \\
    Bicycle           & 0.11      & 15.22      & 18.40  \\
    Pedestrian        & 45.57     & 76.13      & 76.64  \\
    Traffic Cone      & 33.04     & 54.66      & 54.51  \\
    mAP (avg. 10      & \multirow{2}{*}{26.23} & \multirow{2}{*}{46.90} & \multirow{2}{*}{47.92} \\
    classes)          &           &            & \\
    \bottomrule
    \end{tabular}
\label{tab:peformance_nuscenes_16}
\end{table}

\subsection{V2X Collaborative Perception - Experiments}

\subsubsection{Dataset and Metric}
To evaluate our collaboration framework, we use the V2X-Sim 2.0 \cite{li2022v2x-sim} which is made using CARLA \cite{Dosovitskiy17} and the traffic simulator SUMO \cite{SUMO2018}. This dataset is made of 100 100-frame sequences of traffic taking place at intersections of three towns of CARLA which are Town 3, Town 4, and Town 5. Each sequence contains data samples recorded at 5 Hz. Each data sample comprises raw sensory measurements made by the ego vehicle, one to four CAVs, and an IRSU, which is placed at an elevated position that has a large minimally occluded field of view of the intersection. Every vehicle and the IRSU are equipped with a 32-channel LiDAR. All agents are in sync which results in the same timestamp of data that they collect. 

The V2X-Sim 2.0 dataset provides object annotations for each data sample. The official training, validation, and testing split are made of temporally disjoint data samples from three towns chosen such that there is no overlap in terms of intersections. Since we need point cloud sequences as input to our models, we can't use the official splits. Instead, we use sequences in Town 4 and Town 5 as the training set, and those in Town 3 as the validation set, thus ensuring there is no intersection overlap. This choice results in an 8900-data-sample training set and an 1100-data-sample validation set. Since this dataset follows the format of the NuScenes, we use NuScenes' implementation of mean Average Precision (mAP) (details in Sec.\ref{sec:metrics}) to measure the performance of our framework and baselines. 

\subsubsection{Implementation, Experiments and Results}
In the convention of the V2X-Sim dataset, the IRSU and the ego vehicle are respectively assigned the identity of 0 and 1 while other CAVs get identities ranging from 2 to 5. To test our collaboration method's ability to handle asynchronous exchanged information, we set the time lag between the timestamp $t$ of the ego vehicle's query and the timestamp $t_i$ of the detection $\mathcal{B}_i = \left\{\mathbf{b}_{i, j}\right\}$ made by agent $\mathcal{A}_i (i \in \{0, 2, 3, 4, 5\})$ to the time gap between two consecutive data sample of V2X-Sim which is $0.2$ seconds.

In the implementation of our V2X collaboration framework, every agent uses the single-vehicle detection model that is developed in Sec.\ref{sec:aligner++} with the absence of the teacher module. 
The architecture and hyper-parameters are kept unchanged as the experiments of the single-vehicle model in Sec.\ref{sec:single_vehicle_implementation_detail}. 
While previous works using the V2X-Sim dataset \cite{li2021disconet, li2022v2x-sim} set the detection range to $[-32, 32]$ meters along the X and Y axis, centered on the ego vehicle, we extend this range to $[-51.2, 51.2]$ to better demonstrate the performance gain thanks to collaborative perception via V2X. 

\paragraph{Enhance Detection of Visible Objects}
We benchmark our approach to collaborative perception against two extremes of the performance-bandwidth spectrum which are Late and Early Collaboration. Late Collaboration achieves the minimal bandwidth usage by fusing the detection $\mathcal{B}_1 = \left\{\mathbf{b}_{1, j}\right\}_{j=1}^{|\mathcal{B}_1|}$ made by the ego vehicle with the detection made by other agents $\left\{\mathcal{B}_i | i \in \{0, 2, 3, 4, 5\} \right\}$ using Non-Max Suppression. We evaluate this baseline under three settings including
\begin{itemize}
    \item \textit{asynchronous} exchange where the gap between $t_i$ and $t$ is $0.2$ seconds as described above
    \item \textit{asynchronous} exchange with $\mathcal{B}_i$ propagated from $t_i$ to $t$ using scene flow by the procedure described in section \ref{sec:modar_for_v2x}
    \item \textit{synchronous} exchange where there is no gap between $t$ and $t_i$
\end{itemize}
On the other hand, Early Collaboration reaches high performance by exchanging the entire raw point cloud sequences $\mathcal{S}_i = \left\{\mathcal{P}^{t_i - K + 1}_i, \dots, \mathcal{P}^{t_i}_i\right\}$ collected by each agent $\mathcal{A}_i$ is the second baseline. 

\begin{table}[htb]
    \caption{performance of collaboration methods on ground truth that are visible to the ego vehicle}
    \centering
    \begin{tabular}{l c}
    \toprule
    Collaboration Method                & mAP \\
    \midrule
    None                                &  65.71 \\
    
    \midrule
    Late - \textit{async}               &  54.77 \\
    Late - \textit{async} prop.         &  59.34 \\
    Late - \textit{sync}                &  61.29 \\
    
    \midrule
    Early                               & \underline{71.84} \\
    
    \midrule
    Ours                        & \textbf{75.19} \\
    \bottomrule
    \end{tabular}
\label{tab:v2x_performance_ego_gt}
\end{table}

To verify the ability to enhance the single-vehicle perception of V2X collaboration, we evaluate our approach and baselines in the setting where ground truths are made of objects visible to the ego vehicle. 
This means eligible ground truth must contain at least one point of the point cloud $\mathcal{P}_e^t$, which the ego vehicle obtained at the time step of query $t$. 
The result of this evaluation is summarized in Tab.\ref{tab:v2x_performance_ego_gt} which shows that Late Collaboration in all three settings is not beneficial, as the best Late Collaboration is only $93.3\%$ (4.42 mAP behind) of the single-agent perception. 
This is because True Positive (TP) detections made by other agents are counted as false positives if they are not visible to the ego vehicle. 
In addition, ill-localized but overly confident detections made by other agents can suppress good detections made by the ego vehicle, thus further reducing the overall performance. 
Remarkably, our collaboration approach using \textit{MoDAR} points outperforms the early collaboration by $4.7\%$ (3.35 mAP). 
This reaffirms our design philosophy that \textit{a good multi-agent collaborative perception framework can be made on the foundation of a good single-agent perception model and a simple collaboration method}.

\paragraph{Enhance Detection of Completely Obstructed Regions Objects}
Besides enabling the ego vehicle more accurately detect objects that are visible to itself, another great benefit of collaborative perception is to help the ego vehicle see the objects that are completely obstructed - those that do not contain any points of its point cloud $\mathcal{P}_e^t$, thus overcoming the challenge of occlusion and sparsity. 
To demonstrate this, we relax the criterion of eligibility of ground truth such that they only need to contain at least one point emitted by any agents in the V2X network. 
In this experiment, we add a strong mid-collaboration method - DiscoNet \cite{li2021disconet} to the set of baselines to showcase the capability of our approach compared to the state-of-the-art. 
In the framework of DiscoNet, connected agents, which have identical detection architecture, exchange BEV images of their point cloud.
The ego vehicle fuses the BEV image of its point cloud with the exchanged BEV via a fully connected collaboration graph to use as the input to the detection head.

In addition to the detection precision, we also measure the bandwidth consumption of each collaboration method. 
Our measurement is based on the raw uncompressed form of the exchanged data to facilitate a fair comparison. 

The results shown in Tab.\ref{tab:v2x_performance_full_gt} indicate that on a larger set of ground truths, Late Collaboration does improve performance, compared to single-vehicle (no collaboration). 
The improvement is significant (at least $8.35$ mAP) even in the poorest setting where the set of exchanged detection $\left\{\mathcal{B}_i | i \in \{0, 2, 3, 4, 5\} \right\}$ is $0.2$ seconds behind the time of query, thus resulting a spatial misalignment between detected objects that are exchanged and their associated ground truths if the underlying objects are dynamic. 
This can be explained by a significant number of static and slow-moving objects present in intersections whose past detections remain true positives at the present. 
When detections of dynamic objects are perfectly accounted for as in the \textit{sync} setting where agents exchange their detections at the same timestamp as the query, the performance is largely improved by $9.29$ mAP ($15.2\%$), compared to the \textit{async} setting. 
However, this \textit{sync} setting is unrealistic because different agents have different detection rates. 
Interestingly, propagating detections using scene flow as in the \textit{async} prop. setting can reach $96.2\%$ the performance of the \textit{sync} setting ($2.68$ mAP behind). 
This implies the effectiveness of scene flow estimated by our \textit{Aligner++}.

\begin{table}[htb]
    \caption{performance of collaboration methods, measured by mAP, on ground truth that are visible to at least one agent}
    \centering
    \begin{tabular}{l l l r}
    \toprule
    Collaboration                           & \multirow{2}{*}{Sync}     & \multirow{2}{*}{Async}  & Bandwidth \\
    Method                                  &                           &                         & Usage (MB) \\
    \midrule
    None                                    & 52.84                     & -                       & 0    \\

    \midrule
    \multirow{2}{*}{Late}                   & \multirow{2}{*}{70.48}    & 61.19                   & \multirow{2}{*}{\textbf{0.01}}  \\
                                            &                           & 67.80~\footnotemark[1]               & \\

    \midrule
    Mid - DiscoNet~\footnotemark[2]
                                            & \underline{78.70}         & 73.10       & 25.16 \\
    
    \midrule
    Early                                   & 78.10                     & \textbf{77.30}                       & 33.95 \\
    \midrule
    
    Ours                                    & \textbf{79.20}            & \underline{76.72}          & \textbf{0.01}  \\
    
    \bottomrule
    \end{tabular}
\label{tab:v2x_performance_full_gt}
\end{table}

\footnotetext[1]{Exchanged detections are propagated using scene flow according to (\ref{eq:propagate_modar}) before fusion by Non-Max Suppression.}
\footnotetext[2]{Our implementation based on \cite{li2021disconet}}

In the \textit{sync} setting, DiscoNet narrowly exceeds Early Collaboration by $0.6$ mAP ($0.7\%$), thus showing the benefit of the deep interaction among agents powered by their dense collaboration graph.
However, this benefit comes at the cost of relatively high bandwidth usage of $25$ MB per broadcast BEV image on average which is only $25\%$ less than the average bandwidth required by Early Collaboration.
The operation of our method in this setting is as follows: \textit{MoDAR} points broadcast by other agents have the same timestamp as the point cloud of the ego vehicle; therefore their propagation using scene flow is skipped. 
Our method slightly exceeds DiscoNet by $0.5$ mAP ($0.6\%$) to reach the highest performance while consuming the least bandwidth among collaborative methods. 
This confirms the effectiveness of \textit{MoDAR} in facilitating collaboration among connected agents.

In the \textit{async} setting, DiscoNet is severely affected by the misalignment among exchanged BEV images caused by timestamp mismatch as its performance dropped by $5.6$ mAP ($7.1\%$).
Our method is affected because the misalignment between exchanged \textit{MoDAR} points and their corresponding ground truth is explicitly taken into account by the propagation of \textit{MoDAR} points using scene flow.
As a result, the performance gap between ours and DiscoNet is extended to $3.62$ mAP ($5\%$).
Early Collaboration shows the most robustness against timestamp mismatch with an $0.8$ mAP ($1\%$) drop.
This can be explained by the detector's ability to implicitly handle to some extent the misalignment among point clouds.
This phenomenon is well known in the literature on multi-frame single-vehicle object detection \cite{caesar2020nuscenes, zhu2019class, yin2021center, yang20213d} where models trained on the concatenation of 0.5-second point cloud sequences enjoy a $30\%$ mAP gain compared to models trained on individual point clouds.

\begin{figure}[b]
    \centering
    \includegraphics[width=0.98\linewidth]{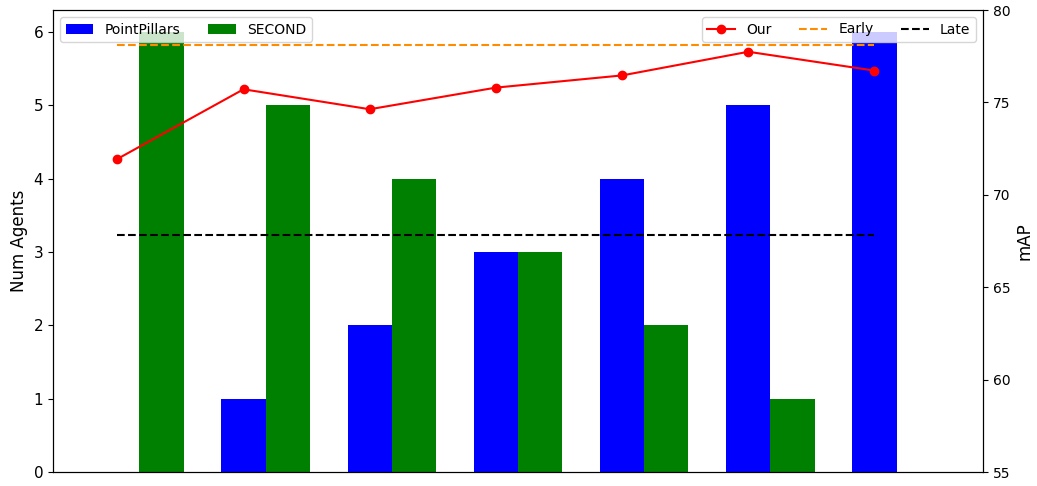}
    \caption{Performance of the ego vehicle in heterogeneous collaboration networks made PointPillar and SECOND detectors. 
    As the number of SECONDs increases from 0 to 6 - the size of the network, the ego vehicle's performance transits from its value in a full-PointPillar network to its value in a full-SECOND network without catastrophe dropping out of the lower bound of Late Collaboration.}
\label{fig:mix_pillar_second_mAP}
\end{figure}

\paragraph{Support Heterogeneous Detector Networks}
To show that our collaboration method supports heterogeneous networks out of the box, we perform an experiment where we start with a network of agents uniformly using PointPillar as their detectors and gradually replace PointPillar with SECOND in one agent after another. 
The result of this experiment is shown in Fig.\ref{fig:mix_pillar_second_mAP}. 
As the number of SECOND increases, the performance of the collaborative perception gradually transits in a downward trend from $76.72$ mAP of a full-PointPillar network to $71.94$ mAP of a full-SECOND network, without dropping below the lower bound of Late Collaboration, which is the case of mid-collaboration method without dedicated modules for bridging the domain gap \cite{xu2022bridging}. 
This downward trend is because an individual PointPillar achieves higher AP than an individual SECOND on the V2X-Sim dataset when being trained under the same setting. 
More interestingly, the combination made of the RSU using SECOND and all connected vehicles using PointPillar archives the best performance of $77.74$ mAP, reaching $99.5\%$ performance of Early Collaboration in \textit{sync} setting.

\begin{figure}[b]
    \centering
    \includegraphics[width=0.98\linewidth]{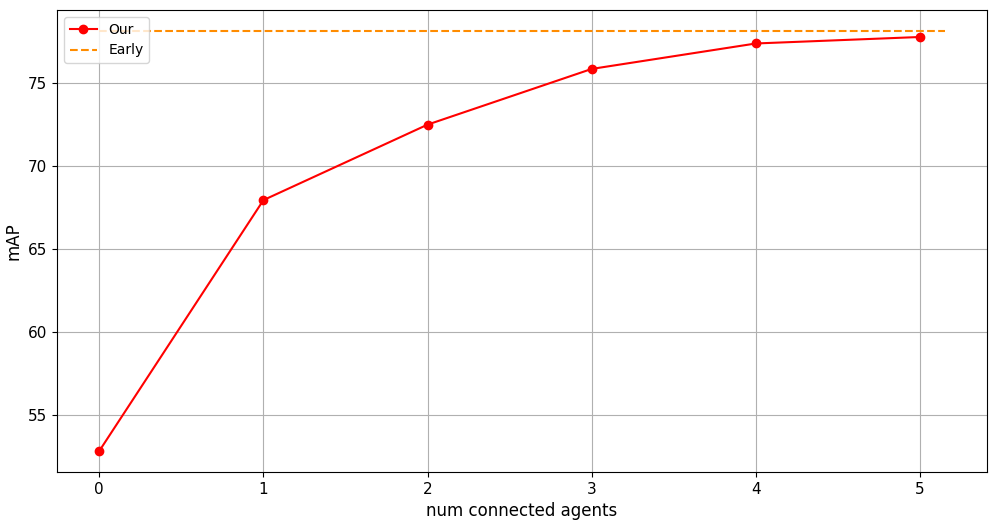}
    \caption{The performance of ego vehicle's detection increases sharply in collaboration with only IRSU, then slows down and saturates at $98.2\%$ of Early Collaboration in \textit{sync} setting as the number of connected agents increases.}
\label{fig:perf_vs_num_agents}
\end{figure}

\paragraph{Relation between Performance and Network Size}
Finally, to study how the performance of collaboration perception evolves with respect to the number of agents in the V2X networks, we gradually increase the number of participants in the collaboration.
As can be seen in Fig.\ref{fig:perf_vs_num_agents}, the collaboration with only the IRSU brings $15.09$ mAP ($28.6\%$) improvement.
The magnitude of performance gain reduces and eventually saturated as the number of agents increases.
An increased number of agents results in higher overlap in agents' field of view which makes objects in the overlapped regions well detected by many agents.  
However, this overlap also means the field of view of the collaboration network as a whole does not increase proportionally with the number of agents.
Moreover, exchanging detected objects made by distant agents may not get used by the ego vehicle because they are outside the predefined detection range.

\begin{figure*}[htb]
    \centering
    \subfloat[\small 1 RSU \& 3 CAVs]{
        \includegraphics[trim={10.5cm 12.5cm 10.5cm 12.5cm},clip,width=.32\linewidth]{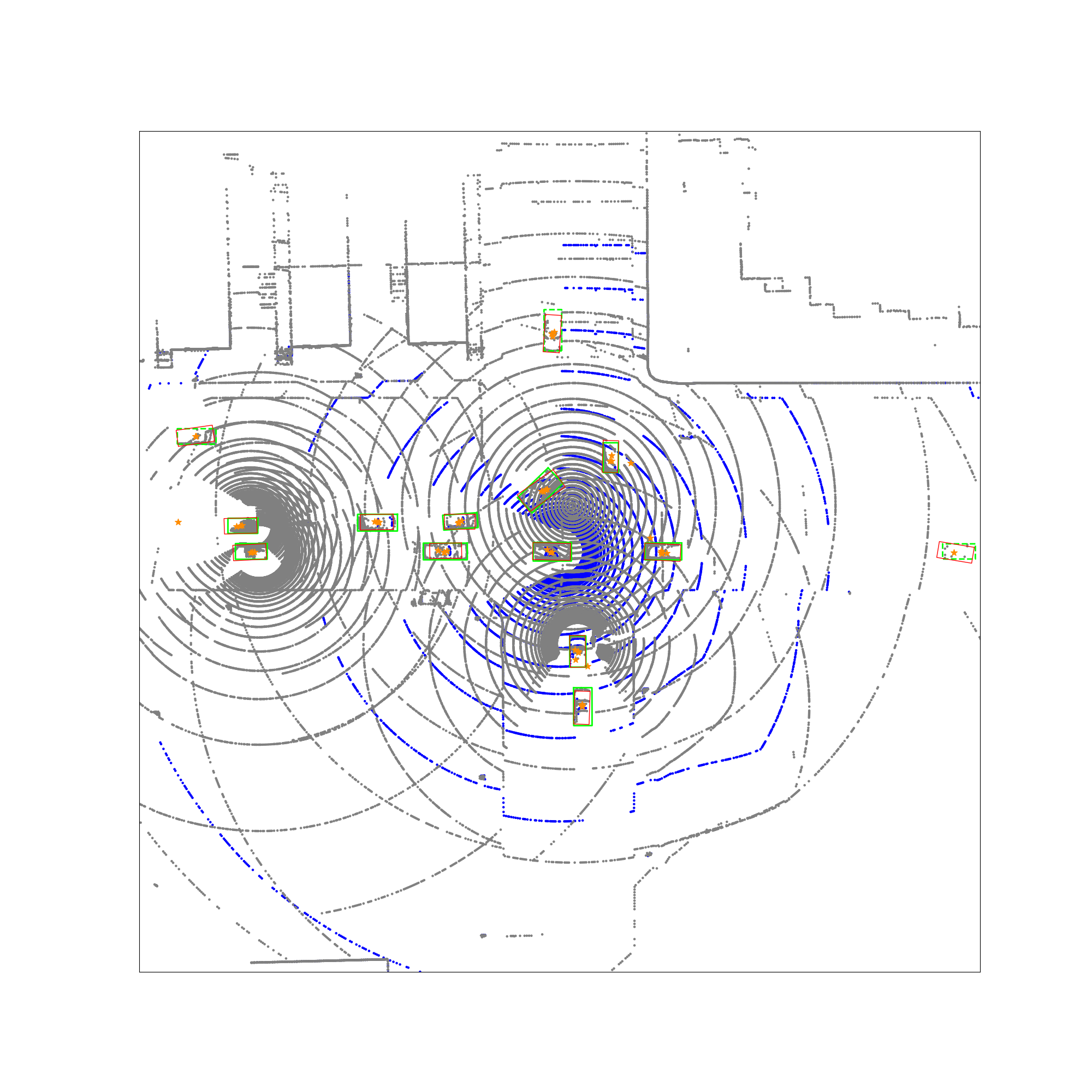}
    }
    \hfil
    \subfloat[\small 1 RSU \& 1 CAVs]{
        \includegraphics[trim={10.5cm 12.5cm 10.5cm 12.5cm},clip,width=.32\linewidth]{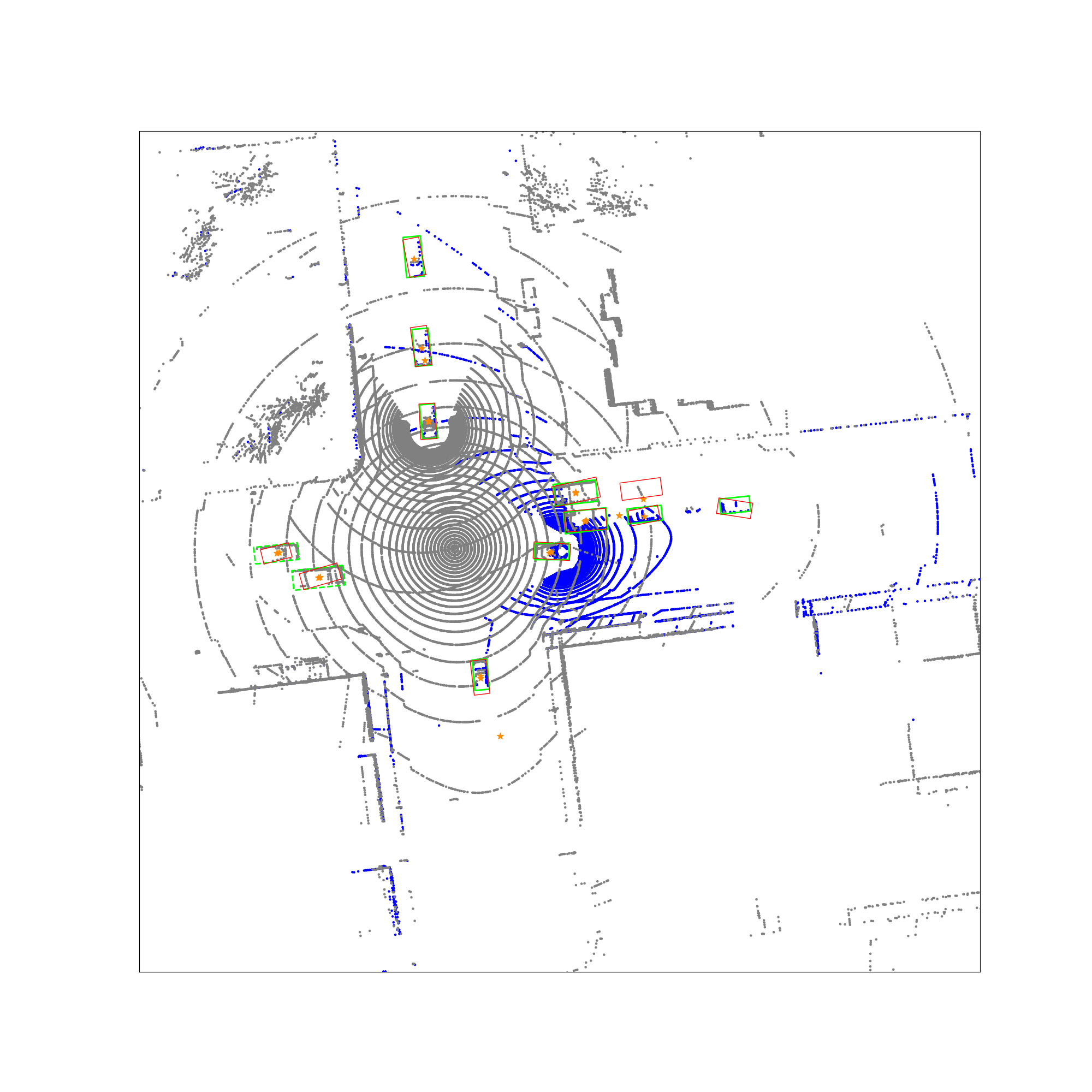}%
    }
    \hfil
    \subfloat[\small 1 RSU \& 3 CAVs]{
        \includegraphics[trim={10.5cm 12.5cm 10.5cm 12.5cm},clip,width=0.32\linewidth]{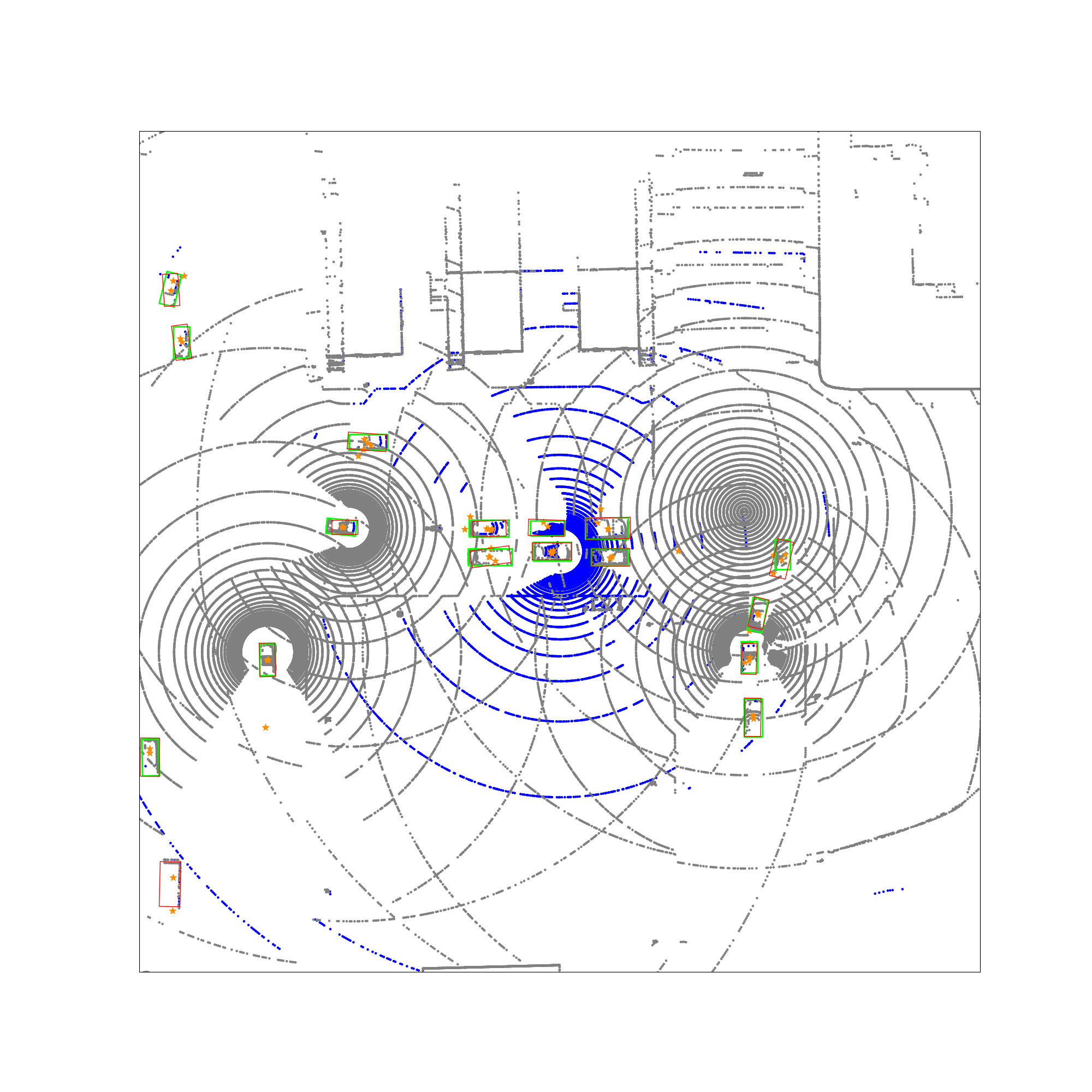}%
    }
    \\
    \subfloat[\small 1 RSU \& 2 CAVs]{
        \includegraphics[trim={10.5cm 12.5cm 10.5cm 12.5cm},clip,width=.32\linewidth]{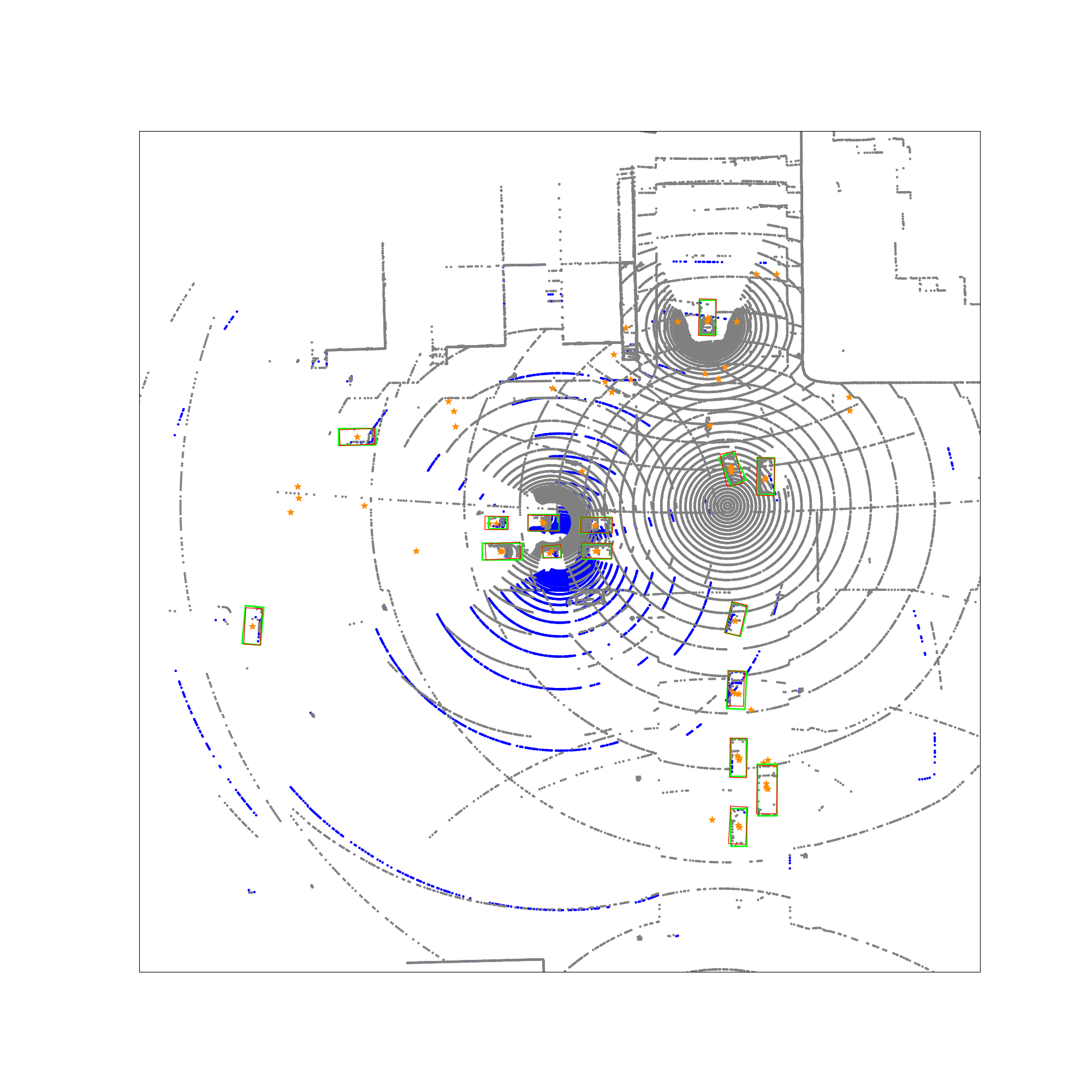}
    }
    \hfil
    \subfloat[\small 1 RSU \& 3 CAVs]{
        \includegraphics[trim={13.5cm 12.5cm 7.5cm 12.5cm},clip,width=.32\linewidth]{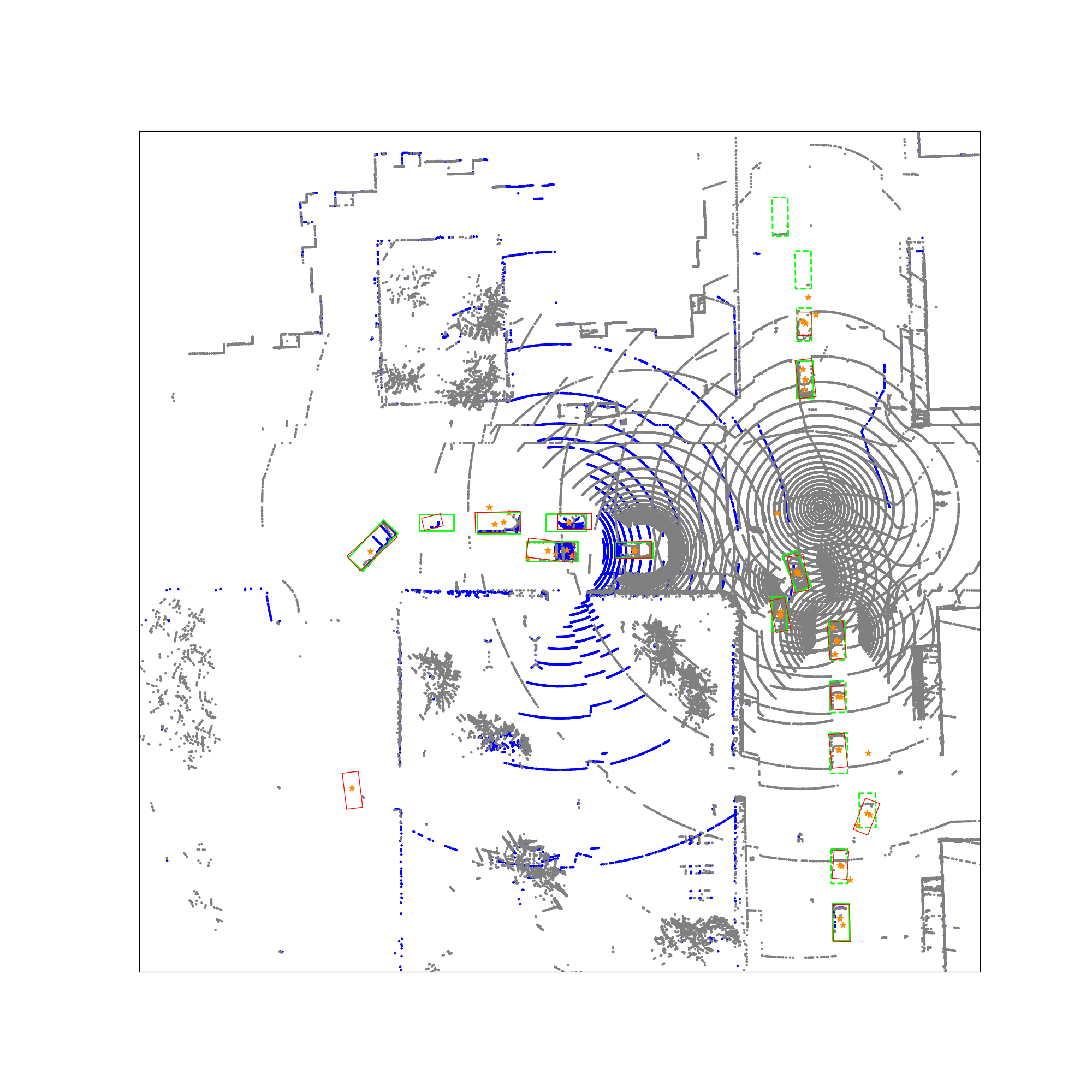}%
    }
    \hfil
    \subfloat[\small 1 RSU \& 1 CAVs]{
        \includegraphics[trim={15.5cm 9.5cm 5.5cm 15.5cm},clip,width=0.32\linewidth]{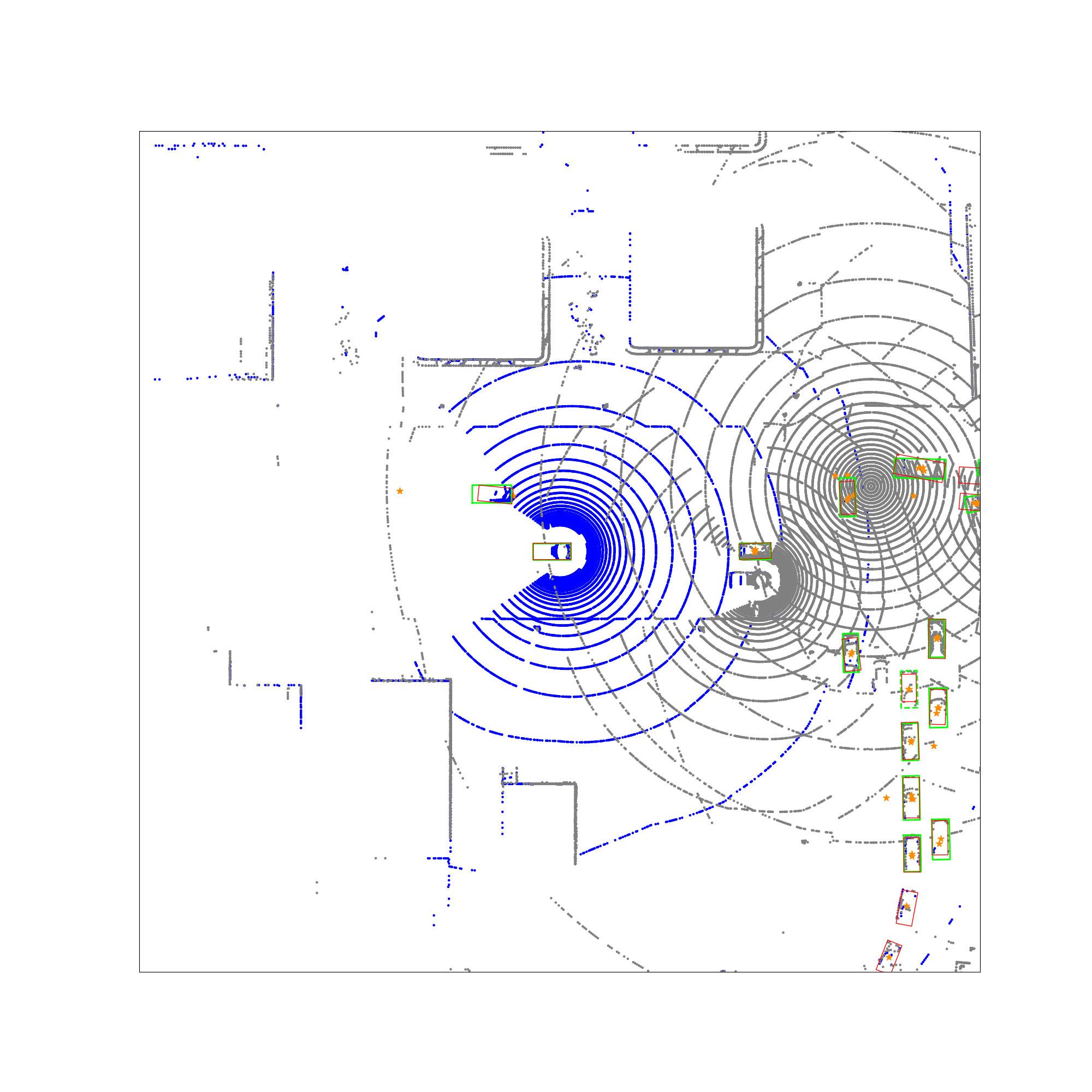}%
    }
    \caption{Qualitative performance of our method on three samples of the V2X-Sim dataset. 
    The number and type of connected agents are annotated in the title of each image.
    \textcolor{blue}{Blue points} are LiDAR points collected by the ego vehicle. 
    \textcolor{gray}{Gray points} are LiDAR points collected by other agents which are displayed for visualization purposes only.
    \textcolor{orange}{Orange stars} denote the MoDAR points broadcast by other connected agents.
    \textcolor{green}{Green solid and dashed rectangles} respectively represent ground truth visible and invisible to the ego vehicle.
    \textcolor{red}{Red rectangles} are the detections made by the ego vehicle using our method.}
\label{fig:qualitative_performance}
\end{figure*}

\paragraph{Qualitative}
The qualitative performance in Fig.\ref{fig:qualitative_performance} shows that collaborative perception using \textit{MoDAR} points enables the ego vehicle to detect objects that are occluded or have a few to zero LiDAR points due to long range.
Particularly, the vehicle at the top of Fig.\ref{fig:qualitative_performance}a is occluded with respect to the ego vehicle due to the presence of a large vehicle. However, this occluded vehicle is successfully detected thanks to a single MoDAR point produced by the IRSU.
Another example of occluded vehicles that are successfully detected is Fig.\ref{fig:qualitative_performance}e where vehicles in the bottom right corner are occluded by a building-like structure.
The illustration of successful detection at long range thanks to \textit{MoDAR} points can be found in Fig.\ref{fig:qualitative_performance}b, Fig.\ref{fig:qualitative_performance}d, and Fig.\ref{fig:qualitative_performance}f where vehicles near their edges have a few or even zero LiDAR points of the ego vehicle.

\section{Conclusions and Future Work}
In this paper, we develop a collaborative perception framework based on the foundation made of \textit{good single-agent perception models} and \textit{an effective late-early collaboration method}.
The simplicity of our collaboration method facilitates its practicality by (i) minimizing bandwidth usage, (ii) dismissing the need for inter-agent synchronization, (iii) making minimal changes to single-agent object detectors, and (iv) supporting networks of heterogeneous detectors.
Importantly, this practicality does not compromise the overall performance as our collaboration method exceeds \textit{Early Collaboration} on the V2X-Sim dataset.
The success here paves the way for our next development which is the demonstration of our method on a real-world V2X network.

\section{Acknowledgement}
This work was granted access to the HPC resources of IDRIS under the allocation 2021-AD011012128R1 made by GENCI. 
This work has been supported in part by the ANR AIby4 under the number ANR-20-THIA-0011. 
This work was carried out in the framework of the NExT Senior Talent Chair DeepCoSLAM, which is funded by the French Government, through the program Investments for the Future managed by the National Agency for Research (ANR-16-IDEX-0007), and with the support of Région Pays de la Loire and Nantes Métropole.


\vfill

\end{document}